\documentclass[11pt]{article}

\usepackage[margin=1in]{geometry}
\usepackage{amsmath,amssymb,amsfonts}
\usepackage{graphicx}
\usepackage{booktabs}
\usepackage[ruled]{algorithm2e}
\usepackage{mdframed}
\usepackage{xcolor}
\usepackage{comment}
\usepackage{subcaption}
\usepackage{threeparttable}
\usepackage{setspace}
\usepackage{booktabs}
\usepackage{multirow}
\usepackage[T1]{fontenc}

\usepackage{parskip}

\usepackage[numbers,sort&compress]{natbib}
\usepackage[colorlinks,citecolor=blue,linkcolor=blue,urlcolor=blue]{hyperref}

\SetAlFnt{\small}
\SetAlCapFnt{\small}
\SetAlCapNameFnt{\small}
\SetAlCapHSkip{0pt}
\IncMargin{-\parindent}

\makeatletter
\g@addto@macro\@floatboxreset\centering
\makeatother

\newcommand{\xhdr}[1]{\vspace{1mm}\noindent{\bf #1}}
\newcommand{\Xtwitter}{\ensuremath{\mathbb{X}}}
\usepackage[]{color-edits}
\addauthor{cp}{red}
\addauthor{am}{blue}
\addauthor{sc}{red}

\newif\ifarxiv
\arxivtrue

\newmdenv[
  linewidth=0.8pt,
  backgroundcolor=gray!5,
  roundcorner=3pt,
  innerleftmargin=8pt,
  innerrightmargin=8pt,
  innertopmargin=8pt,
  innerbottommargin=8pt,
  skipabove=\baselineskip,
  skipbelow=\baselineskip
]{promptbox}

\title{Using AI Agents to Automate Black-Box Audits\\of Personalization Algorithms at Scale}

\author{%
    Alessandro Morosini$^{*,1}$\quad
    Sarah H.\ Cen$^{*,2}$\quad
    Andrew Ilyas$^{2}$\\[4pt]
    Hedi Driss$^{1}$\quad
    Aleksander M\k{a}dry$^{1}$\quad
    Chara Podimata$^{1}$\\[8pt]
    $^{1}$Massachusetts Institute of Technology\\
    $^{2}$Carnegie Mellon University
}

\newcommand{\affilblock}{%
  \begingroup
    \renewcommand{\thefootnote}{}%
    \footnotetext{\textsuperscript{*}Equal contribution.}%
    \addtocounter{footnote}{-1}%
  \endgroup

  \begingroup
    \renewcommand{\thefootnote}{}%
    \footnotetext{%
      Correspondence to:
      \texttt{morosini@mit.edu},
      \texttt{sarahcen@andrew.cmu.edu},
      \texttt{andrewi@andrew.cmu.edu},
      \texttt{hdriss@mit.edu},
      \texttt{madry@mit.edu},
      \texttt{podimata@mit.edu}.
    }%
    \addtocounter{footnote}{-1}%
  \endgroup

  \setcounter{footnote}{0}%
}

\date{\today}

\begin{document}

\maketitle
\affilblock

\begin{abstract}

Personalization algorithms determine what content users encounter on platforms such as social media, search engines, and chatbots. Auditing these systems is difficult because independent auditors typically have only black-box access to the algorithms, while personalization depends on users' attributes, behavior, and evolving interaction histories. Existing auditing methods face the following tradeoff: studies with real users capture realistic behavior but are costly and hard to control, whereas sock-puppet audits scale more easily but often rely on scripted behavior that limits realism. Beyond this tradeoff, both approaches struggle to decouple user attributes from user behavior, limiting our ability to causally understand personalization.

To address this gap, we introduce a framework for black-box audits of personalization algorithms using generative AI agents as behavioral engines for synthetic accounts. Each agent is instantiated with a fixed persona, grounded in demographic and political survey data, and interacts with a platform's content by reasoning about it and choosing actions. Because behavior is fixed within each persona, while platform-visible signals such as age, gender, or location can be experimentally perturbed, our design enables counterfactual auditing of how platforms respond to user attributes under controlled behavioral policies.

As a case study, we deploy 1,120 agents on $\Xtwitter$ shortly after the 2024 U.S. election period, spanning 14 personas and three counterfactual conditions, and collect over 200,000 content exposures. We find that $\Xtwitter$'s algorithmic feed amplifies toxic, polarizing, political, and right-leaning content relative to the chronological feed, with amplification varying sharply by user ideology. Counterfactual analyses further show that demographic signals affect content delivery in heterogeneous, persona-dependent ways: pooled effects are largely null, while subgroup-level effects vary in direction and magnitude. Our work establishes GenAI-based agents as a new methodological tool for algorithmic auditing.

\end{abstract}

\section{Introduction}
\label{sec:intro}

\emph{Personalization} underlies countless digital platforms, including social media, chatbots, and search engines. 
These systems use algorithms to tailor content, responses, and recommendations to users based on their attributes, preferences, and past behavior, thus
shaping interactions and information online, influencing public discourse and social norms while amplifying certain cultural and political narratives over others. Underscoring the growing public and regulatory concern over how algorithmic systems shape what people see, believe, and do online, Congress has convened multiple hearings to question the CEOs of Meta, Amazon, Google, and OpenAI, among others~\citep{house2021misinformation,senate2023bigtech,senate2023aioversight}.

Yet, it has been difficult to scrutinize and audit any kind of algorithmic system at scale, as they are closely guarded as trade secrets; 
the ways in which they are designed and trained, along with the data and user logs associated with them, are often considered proprietary business information. 
Despite repeated calls for external audits by regulators and even tech leaders~\citep{nytimes2023altmanregulation}, 
access to algorithmic internals is rarely granted~\citep{pasquale2015blackbox,brennan2024metaalgorithms}. 
As a result, independent and third-party auditors typically have no more than \emph{black-box access}: the ability to test algorithms on a predetermined set of inputs and observe the outputs \cite{metaxa2021auditing,cen2024transparency}.
This lack of access makes it difficult to establish strong claims about an algorithmic system \cite{casper2024black}, though there are some systematic ways to design black-box experiments \cite{cen2024transparency}. 

Personalization makes this challenge especially severe. Unlike e.g., a static classifier, whose behavior can be probed on a fixed set of inputs, a personalization system responds to a user's attributes and to an evolving interaction history. That history is itself shaped by previous recommendations, creating a feedback loop between the user and the algorithm. As a result, the relevant input space is enormous and composed of all possible user trajectories arising from different attributes, behaviors, exposure histories, and platform responses over time. Exhaustively exploring this space is infeasible, particularly under black-box access. Motivated by these challenges, our work seeks to answer the following question: \emph{how can personalization algorithms be audited at scale with only black-box access?}

Our work answers this question using {\bf LLM-based agents} to demonstrate how personalization algorithms can be audited at scale, despite limited access. 

To situate our approach, 
we contrast it with prior work. 
Historically, one way to audit personalization with black-box access is to recruit human users and observe how the algorithm behaves for them~\citep{bouchaud2023crowdsourced,Wang2024}.
This approach has strong ecological validity---both user behavior and algorithmic responses are real---but it is less suitable for auditing at scale for three main reasons.
First, recruiting human participants is costly. Second, and more fundamentally, researchers have limited control over which user types are represented and thus cannot perform arbitrary causal analyses (e.g., cannot test how the algorithm behaves for users who are not represented by the recruited pool).
Third, humans have pre-existing biases, preferences, and exposures that auditors cannot fully observe or control, making it difficult to isolate algorithmic effects from user-specific factors.

To address the limitations of human studies, 
some researchers use ``sock puppets'', i.e., user accounts controlled by the researchers themselves.
Sock puppets allow more controlled causal analyses and are less costly to scale than human studies, making this approach a powerful tool.
That said, sock puppets can be \emph{limited in their behavioral diversity} and in their \emph{ability to mimic real-world interactions}. 
Prior work controls sock puppets using computer programs that follow predefined rules~\citep{Bandy2021,Haroon2023,Ye2025}. 
The resulting sock puppets can be ``flat'', exhibiting relatively predictable behavior that compromises the applicability of their findings.

Our paper fills this gap by presenting a \textbf{framework for performing black-box audits of personalization algorithms at scale using AI agents}. 
Specifically, we propose to run sock puppet audits powered by AI agents on the target platform of interest.
In contrast to prior work, the actions taken by these synthetic users are controlled by an agent whose behavior is determined by a detailed ``persona'' description that differs across users. Unlike traditional sock-puppet accounts, which typically follow fixed interaction rules, AI agents exhibit richer behaviors. As such, they enable researchers to more comprehensively explore the input space of user conditions to which personalization algorithms tailor content. %
By systematically varying user attributes, personas, and other experiment conditions, 
researchers using our framework are able to run rich counterfactual and comparative analyses, such as: 
``How would the algorithm have behaved if the user had been older?'', ``How does the content differ for users with different political personas?'', 
and ``How do different users' algorithmically personalized feeds differ from their reverse-chronological ones?''.

\subsection{Our Contributions}
\xhdr{Black-Box Auditing Framework.}
We formalize black-box auditing of personalization systems as the identification of \emph{policy-conditional} signal responses, and introduce a framework for black-box auditing using LLM-powered agents. Each agent executes a fixed behavioral policy encoded in a prompt that describes its \emph{persona}. Each persona is characterized by demographic and political ideology attributes, grounded in U.S. Census data~\citep{uscensus2020urban} and Pew's Political Typology~\citep{pew2021typology}. Crucially, both the attributes and the behavior are defined \emph{a priori}, not shaped by platform interaction.
This has two consequences. First, because user attributes such as ideology are specified before any engagement with the platform, researchers can stratify algorithmic outcomes by user characteristics, usually confounded in human-based audits and less realistic with scripted bots. Second, because the behavioral policy is held constant across accounts within a persona, auditors can randomize platform-visible ``signals'' (attributes that a platform can either observe directly or infer confidently, e.g. a user's age) while keeping agent behavior unchanged: this enables causal identification of the platform's response to demographic attributes. Our methodology is general and can be adapted to a broad class of algorithmic auditing settings.
 
\xhdr{Field Experiment on {\Xtwitter}.}
We present the first large-scale deployment of LLM-powered agents for auditing on a live social media platform. We deploy 1,120 agents on {\Xtwitter}\footnote{Our team obtained an IRB exception for the experiment. Our LLM agents only engage with existing content and do not post original content. This minimizes impact on other users' experience.}\footnote{A note on the legality of scraping from {\Xtwitter}: A 2018 court decision (see \url{https://www.aclu.org/documents/sandvig-v-barr-memorandum-opinion}) allowed researchers to violate platforms' terms of service for research purposes. Our team also worked closely with a law clinic to verify the legality of our approach.} over the 24 days following the 2024 U.S.\ presidential election
(November 5--28), a period of heightened political salience. Each agent is grounded in one of 14 personas spanning a range of demographic and ideological profiles. This yields 80 accounts per persona randomly assigned across four conditions (one baseline and three single-attribute counterfactuals over age, gender, and location). Each account browses both the algorithmically curated \emph{For You} feed and the
reverse-chronological \emph{Following} feed. The divergence between these two feeds has attracted growing attention from researchers and the
press alike~\citep{Huszr2022, washpost2023foryou, wsj2024xalgorithm}. In each session, agents observe recommended posts and choose among liking, following the author, reading replies, or ignoring. All these actions are selected by the LLM based on the post's content and the agent's persona. %
The deployment yields over $200{,}000$ account-post exposures in total. %

\xhdr{Evidence of Differential Treatment on {\Xtwitter}.}
 On the aggregate user level, we find that {\Xtwitter}'s recommendation algorithm systematically amplifies toxic, polarizing, and right-leaning content in the \emph{For You} feed relative to the \emph{Following} feed, while left-leaning content is not significantly amplified. Stratifying by user ideology reveals even sharper ideological asymmetries: for example, right-leaning content is amplified for both user groups, while left-leaning content is actively suppressed for right-leaning users. On top of this, our counterfactual design provides causal evidence that the algorithm responds to demographic signals, but not uniformly:
 pooled effects are largely null, yet joint tests reject homogeneity for all outcomes, and persona-level significant effects are roughly double what chance alone would predict. The platform's response to demographic attributes is thus highly persona-dependent: the same signal perturbation can amplify or de-amplify content depending on the user type. Such heterogeneity would be missed by an aggregate study, even one that successfully decouples behavior from signals.

The rest of the paper is organized as follows. Sec.~\ref{sec:related_work} reviews related work. 
Sec.~\ref{sec:setting} formalizes the auditing problem and defines the target estimand. Sec.~\ref{sec:methodology} describes the experimental design and deployment. Sec.~\ref{sec:estimation} presents the measurement and estimation framework. 
Sec.~\ref{sec:results} reports our findings. 
Sec.~\ref{sec:discussion} discusses future directions and the implications of our work.

\subsection{Related Work}
\label{sec:related_work}
\xhdr{Algorithmic Auditing.}
A broad literature has highlighted the societal importance of algorithmic content curation and personalization on search platforms and social media, given their influence on exposure to information \citep{gerhart2004web, bakshy2015exposure, introna2000shaping, elkin2000let, Perreault2024}. 
This influence extends to high-stakes domains such as political elections, where algorithmic curation may shape voter information environments \citep{metaxa2019search, Mustafaraj2020, zade2022auditing, Ye2025}. Algorithmic auditing has emerged as a key methodology for evaluating personalization systems, particularly when internal models and data are unavailable. However, most real-world audits operate under black-box access due to legal limitations and a lack of platform cooperation \citep{sandvig2014auditing}, raising challenges including limited observability, difficulties constructing representative samples, and risks of
biased inference \citep{casper2024black, bandy2021problematic}. This has motivated calls for principled auditing frameworks that improve transparency, reproducibility, and robustness \citep{Imana2023, raji2020closing}.

\xhdr{Human-Based Audits.}
Multiple studies have examined recommender systems using audits that rely on real users, with varying degrees of experimental control. 
Platform-run randomized experiments offer the strongest identification but are rare and subject to platform discretion; examples include {\Xtwitter}'s A/B
test comparing algorithmic and chronological feeds \citep{Huszr2022} and Meta's field experiments that manipulated exposure to algorithmic curation \citep{guess2023social,gonzalez2023asymmetric, nyhan2023like}.
Without platform cooperation, researchers have recruited participants into quasi-experimental designs, such as switchback experiments \citep{Wang2024} or browser-based interventions \citep{piccardi2025reranking}. 
Alternative approaches rely purely on observational data collected via scraping \citep{ribeiro2020auditing} or from consenting users \citep{bouchaud2023crowdsourced,chen2023subscriptions}. 
In general, human-based audits are difficult to scale and do not allow auditors to construct counterfactuals over user profiles or exposure histories, nor to stratify outcomes by latent user characteristics that are endogenous to platform use.

\xhdr{Sock-Puppet Audits.}
More recent work focuses on sock-puppet audits, in which researchers deploy synthetic user accounts to study recommender system behavior. Existing approaches fall into two main categories.
In the first, puppets are trained to a user profile but remain passive, avoiding engagement actions: \citet{Bartley2021} and \citet{Ye2025} define the political orientation of a puppet through curated follow networks, while \citet{Haroon2023} seed puppets with ideological watch histories and track subsequent recommendations.
In the second category, puppets can take actions, but the setting is generally unrealistic: \citet{boeker2022empirical} and \citet{srba2023auditing}
deploy agents on TikTok and YouTube that navigate content via scripted sequences; \citet{Bartley2024} and \citet{chen2021neutral} configure bots to perform simple interactions; other works replace scripted behavior with LLM agents, but deploy them in researcher-built \emph{simulated} environments \citep{yang2411oasis, tornberg2023simulating, ferraro2024agent, wang2025user}.
Instead, our agents reason about each item \emph{and} act on a live platform under black-box access. We further decouple user attributes from behavior, enabling counterfactual auditing of attribute-specific platform responses.

\xhdr{Silicon Sampling.}
A parallel line of research examines whether LLMs can approximate human behavior in experimental settings, sometimes referred to as ``silicon sampling'' \citep{Argyle2022} or ``homo silicus'' \citep{Horton2023}. \citet{Argyle2022} show that GPT-3, when conditioned on demographic backstories, can approximate human opinion distributions across political and social questions. \citet{Horton2023} demonstrate that LLMs can replicate classic behavioral findings at low cost and with flexibility in varying experimental conditions. 
Other work has tested LLM behavior in economic games \citep{Aher2023, Mei2024} and examined behavioral biases %
\citep{leng2024, leng2025}; employed LLMs for market research \citep{Brand2024}; and proposed frameworks for accelerating social science research \citep{manning2024automated, Tranchero2024, hewitt2024predicting}. LLM agents have also been embedded in synthetic social networks to reproduce emergent phenomena such as opinion dynamics and homophily \citep{park2023generative, chang2025llms, gao2023s3}; see \citet{Anthis2025} for a survey.
While a growing literature documents limitations in the ecological validity of LLM-simulated experiments \citep{gui2023challenge, gao2025take, bisbee2024synthetic}, our design mitigates these concerns: personas are defined a priori from survey data rather than generated by the model \citep{Argyle2022, Li2025}, and the LLM serves as a fixed behavioral policy whose potential biases cancel out under within-persona differencing.

\section{Problem Setting and Notation}
\label{sec:setting}
We model the platform as an unknown allocation rule that maps user characteristics and interaction history to recommended content \citep{cen2024measuring}. We aim to isolate how the platform's content allocation changes when \emph{platform-visible user signals} (e.g., declared profile fields and other attributes the platform can infer) are modified, holding the user's \emph{behavior} fixed. We refer to this as \emph{counterfactual auditing}. 

We consider a platform that serves content to a user over a finite time horizon $T$, which may correspond to a fixed deployment window or a number of total interactions. 
Let $\mathcal{Z}$ denote the space of content items (e.g., posts), and let $\mathcal{B}$ denote the set of user actions (e.g., like or follow). Each user account $i \in [N]$ is associated with a time-invariant signal vector $S_i \in \mathcal{S}$ representing user attributes visible to or inferrable by the platform (e.g., user's declared age). 
At each time step $t \in [T]$, the platform serves content $
Z_{i,t} \in \mathcal{Z}$ and the user responds with an action $B_{i,t} \in \mathcal{B}$. Let $H_{i,t} = \{(Z_{i,1},B_{i,1}),\dots,(Z_{i,t},B_{i,t})\}$ denote the interaction history for user account $i$ up to time $t$.

We model the recommendation algorithm as an unknown black-box mapping
$
    \mathcal{A}: (S_i, H_{i,t-1}) \mapsto \mathcal{D}(\mathcal{Z}),
$
where $\mathcal{D}(\mathcal{Z})$ denotes a distribution over content items. 
User behavior is generated by a \emph{policy} $\pi$: a complete behavioral specification that encodes the user's full type, including both demographic characteristics and latent attributes such as political ideology or education level (in our implementation, $\pi$ corresponds to the LLM prompt). %
Given a certain content item $Z_{i,t}$, the policy induces a distribution over actions:
$
B_{i,t} \sim \pi(Z_{i,t}).
$
The auditor summarizes each trajectory via an outcome $Y_i = g(Z_{i,1:T}, B_{i,1:T})$. For example, $g$ may capture properties of the content a user is exposed to that are commonly studied in auditing, such as its average toxicity, the share that is political, and the balance of left- versus right-leaning content. %

\xhdr{Policy-Conditional Signal Response.} Fix a user policy $\pi$ and consider two different values $s_0, s_1 \in \mathcal{S}$ of a platform-visible signal (e.g., $\pi$ might encode a progressive user, while $s_0$ and $s_1$ correspond to profiles displaying ages 25 and 55 to the platform, respectively). %
We define the policy-conditional signal response as the difference in expected outcomes when \emph{the same policy} $\pi$ is paired with different signals:
\begin{equation}
  \Delta_{\pi}(s_1,s_0)
  \;=\;
  \mathbb{E}\!\left[ Y \mid S=s_1,\pi \right]
  \;-\;
  \mathbb{E}\!\left[ Y \mid S=s_0,\pi \right],
  \label{eq:signal_response}
\end{equation}
where expectations are over the distribution of histories induced by $(\mathcal{A}, \pi, s_1)$ and $(\mathcal{A}, \pi, s_0)$, respectively. 
Conceptually, $\Delta_\pi$ measures how the \emph{expected outcome differs} (e.g., the share of toxic content) %
if the \emph{same behavioral policy} were exposed to the platform under \emph{different signal} assignments.

Measuring $\Delta_\pi$ in Eq.~\eqref{eq:signal_response} via black-box access is difficult for two reasons. %
First, black-box access precludes counterfactual simulation: the
auditor cannot re-query the algorithm under modified inputs and
must instead observe realized trajectories from deployed accounts.
Second, even if counterfactuals were observable, existing audit designs fail to decouple \emph{behavior} from \emph{platform-visible signals}. As a result, current audits compare users that differ simultaneously in both, estimating
  \begin{equation}
  \mathbb{E}[Y \mid S=s_1, \pi=\pi_1] \;-\; \mathbb{E}[Y \mid S=s_0, \pi=\pi_0],
  \label{eq:confounded_contrast}
  \end{equation}
where $\pi_1 \neq \pi_0$, rather than the policy-conditional signal response that holds $\pi$ fixed. 
This coupling arises naturally in human audits because for an individual, the behavior (policy) and its induced histories are correlated with $S$.
It also arises in sock-puppet audits because intended ``types'' are usually instantiated through scripted behaviors, where the policy explicitly depends on $S$.

The comparison in Eq.~\eqref{eq:confounded_contrast} is not invalid, but it conflates the platform's response to user signals with its response to user behavior, limiting our causal understanding of personalization systems.
Addressing these limitations requires a design that decouples \emph{behavior} from \emph{signals} and enables counterfactual comparison across different signal assignments; we present such a framework next.

\section{Experiment Design}
\label{sec:methodology}
Our experimental design centers on deploying LLM-powered agents that interact with the platform's recommendation system in a closed feedback loop. Unlike common sock puppets, which follow scripted rules of engagement, our agents \emph{reason} about each piece of content before deciding how to act. Each agent is instantiated with a fixed prompt, a \emph{persona}, that encodes both demographic attributes (e.g., age, gender, location, race) and political ideology. Given this persona, the LLM-powered agent determines autonomously how to respond to each piece of content it encounters, with no further intervention from us. This produces a realistic feedback loop between user actions and platform recommendations, where behavior follows from a fixed \emph{identity} rather than hard-coded rules. In the language of Section~\ref{sec:setting}, the persona prompt captures the user's policy $\pi$.

\begin{figure}[htbp]
    \centering
    \includegraphics[width=\textwidth]{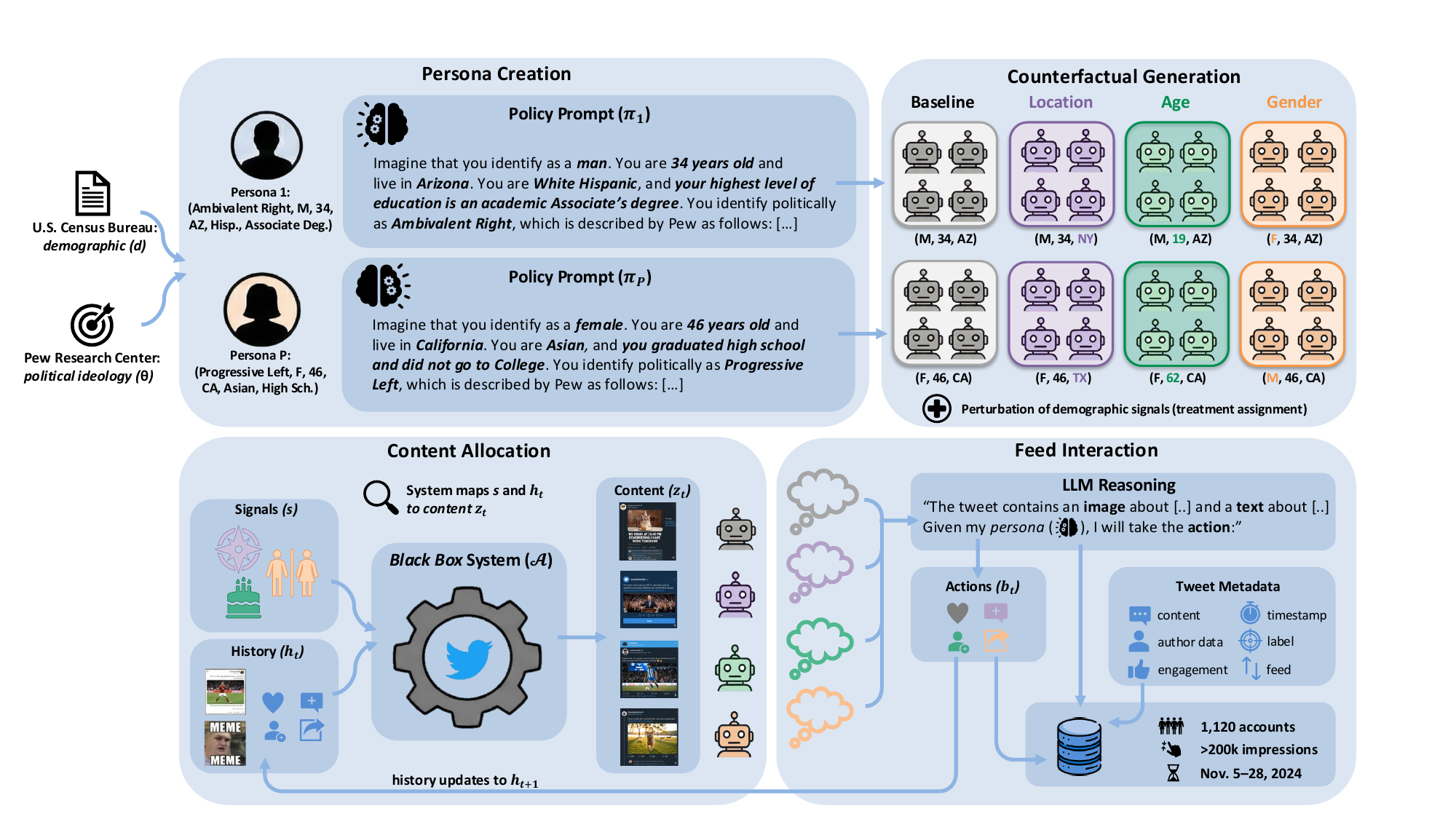}
    \caption{Overview of our experimental framework. \textbf{Top:} each persona is constructed by combining demographics from the U.S.\ Census with a political typology from Pew Research, producing an LLM prompt that fully specifies the agent's behavior policy $\pi_p$. Each persona is replicated into multiple accounts that share the policy but differ in a single platform-visible signal (location, age, or gender) relative to the baseline.  
    \textbf{Bottom:} The interaction loop for a single account. The platform's black-box recommendation system $\mathcal{A}$ observes the account's signal $s$ and interaction history $h_t$ to allocate content $z_t$. The replicas reason about content via the same policy, and select an action $b_t$; actions are appended to form the updated history $h_{t+1}$, creating a feedback loop between agents and platform recommendations. All interactions, reasoning traces and metadata are logged.}
    \label{fig:pipeline}
\end{figure}

Because the behavior policy is held constant across replicas of the same persona, the design supports within-persona randomization of platform-visible demographic signals. This allows us to isolate the causal effect of attributes such as age, gender, or location on algorithmic recommendations, the policy-conditional signal response defined in Equation~\eqref{eq:signal_response}. Our methodology is general and applies whenever (i) a platform serves a sequence of content items to an account, (ii) recommendations depend on platform-visible account signals in addition to interaction history, and (iii) an auditor can deploy multiple accounts and log realized outcomes. An overview of the framework is presented in Figure \ref{fig:pipeline}.

\subsection{Overview}
\label{sec:overview}

We introduce \emph{personas}, indexed by $p \in [P]$, each associated with a fixed behavior policy $\pi_p$. Accounts are indexed by $i \in [N]$ and we let $\mathcal{I}_p$ be the set of users associated with persona $p$. Each account $i \in \mathcal{I}_p$ follows the corresponding policy $\pi_p$. Persona creation details are presented in Section~\ref{sec:persona_and_counterfactuals}. Since our accounts share the same policy within each persona (and hence also the same behavior), we also refer to these accounts as \emph{replicas}. All replicas interact with the platform following the same protocol, as described in Section~\ref{sec:platform_interaction}. 

Because each persona's policy $\pi_p$ is fixed by construction and encodes latent characteristics (e.g., political ideology) \emph{a priori}, the design enables stratifying any outcome by user type without the endogeneity that plagues observational studies. In observational settings, any proxy for a user's ideology derived from engagement history is itself shaped by algorithmic recommendations through the feedback loop between $\mathcal{A}$ and $H_{i,t}$, making the stratification variable a joint product of the user and the algorithm. As our agents' policies are specified before any platform interaction, this conditioning is well-defined by construction. In our $\Xtwitter$ case study, we exploit this to measure policy-conditional amplification of content in the \emph{For You} feed compared to the \emph{Following} feed.
         
Additionally, replicas differ only through a \emph{profile condition} assignment $C_i \in \{0,1,\dots,K\}$, where $C_i=0$ denotes a baseline condition for user $i$ and each $C_i=k$ (for $k\ge 1$) denotes a perturbation to a platform-visible signal. Within each persona $p$, we assign $C_i$ by balanced randomization: replicas in $\mathcal{I}_p$ are split into $K + 1$ equal-size groups, one per condition. In our deployment, we use $P=14$ personas and $|\mathcal{I}_p|=80$ replicas per persona, with $K=3$ counterfactual conditions plus the baseline, splitting replicas into four equal-size groups of $20$ accounts each.

For each counterfactual condition $k$, our target estimand is the policy-conditional average treatment effect relative to baseline,
\begin{equation}
\tau_{p,k}
\;=\;
\mathbb{E}\!\left[\,Y \mid C=k,\ \pi_p \right]
\;-\;
\mathbb{E}\!\left[\,Y \mid C=0,\ \pi_p \right],
\label{eq:ate_persona}
\end{equation}
where $Y$ is a metric of interest, e.g., the content toxicity score or its political leaning; see Section~\ref{sec:metrics} for more details. 
Within each persona, each account's condition assignment $C_i$ determines which component of the platform-visible signal $S_i$ is perturbed relative to the baseline, while the behavior policy is held fixed at $\pi_p$ by construction. The estimand $\tau_{p,k}$ therefore takes the form of a policy-conditional signal response $\Delta_{\pi_p}(s_1, s_0)$ as in Eq. \eqref{eq:signal_response}, comparing the signal induced by condition $k$ against the baseline. In Section~\ref{sec:counterfactual}, we present a regression model for estimating this quantity from the experimental outcomes.

\subsection{Personas and Counterfactuals} 
\label{sec:persona_and_counterfactuals}
We now describe how we construct personas and assign counterfactual conditions. The key design requirement is that all replicas within a persona share an identical behavior policy $\pi_p$, while differing only in their platform-visible signal $S_i$, as determined by the condition assignment $C_i$. 

\xhdr{Persona Construction.}
Each persona $p$ is defined by a tuple of attributes $(\theta_p, d_p)$, where $\theta_p \in \Theta$ denotes a political typology and $d_p = (\text{age}_p, \text{gender}_p, \text{location}_p, \text{race}_p, \text{education}_p)$ denotes demographic characteristics. We use data from the U.S.\ Census Bureau \cite{uscensus2021demographics} (for demographic attributes) and Pew Research Center \cite{pew2021typology} (for political typology) to ground these attributes in real-world population distributions.

Demographic attributes are sampled to approximate the distribution of platform users. %
Age is drawn from four brackets (18--29, 30--49, 50--64, 65+) according to Census population shares, weighted by age-specific platform usage rates from Pew's social media surveys. 
Gender is sampled according to Census proportions (male, female, or other). 
Race and ethnicity follow Census proportions, collapsed to seven categories (White Non-Hispanic, White Hispanic, Black, Asian, American Indian, Native Hawaiian/Pacific Islander, and Multiracial). %
Location is drawn from U.S.\ cities with population exceeding 50,000, with sampling probability proportional to city population. 
Finally, education is sampled conditional on age, gender, and race using Census educational attainment tables. 
Exact sampling weights are reported in Appendix~\ref{sec:demographics}.

For political typology, we use Pew's nine-category Political Typology \citep{pew2021typology}, which partitions the U.S.\ electorate based on political values and party affiliation. These categories span the entire range of political ideologies and are rather fine-grained. Presenting them from right-most to left-most leaning, the categories are: %
Faith and Flag Conservatives, Committed Conservatives, Populist Right, Ambivalent Right, Stressed Sideliners, Outsider Left, Democratic Mainstays, Establishment Liberals, and Progressive Left. More details on how the typology works and what the characteristics of each type are can be found in Appendix~\ref{sec:pol}. We sample $\theta_p$ %
conditional on the persona's age and gender, using Pew's published demographic breakdowns of each typology.

\xhdr{Behavior Policy.}
Each persona $p$ is associated with a behavior policy $\pi_p: \mathcal{Z} \to \mathcal{D}(\mathcal{B})$ mapping observed content to a distribution over actions. We implement $\pi_p$ using GPT-4o with a fixed prompt that encodes the persona's attributes in natural language. More detail on the prompt can be found in Appendix~\ref{sec:policy}.  
Concretely, when the agent encounters some content item $z \in \mathcal{Z}$, the item is sent to GPT-4o together with the persona prompt. The model then returns an action $b \in \mathcal{B}$ based on the encoded identity, with temperature set to zero. The full interaction protocol is presented in Section~\ref{sec:platform_interaction}. 

Each account is also assigned platform-required identifiers needed for deployment. Profile images depict non-human subjects (e.g., landscapes) and 
are generated using DALL-E \citep{ramesh2021zero}, carrying no demographic signal. Account emails are created via MailSlurp, a disposable-inbox service, and usernames are decided by {\Xtwitter} at registration. These identifiers are held fixed throughout the experiment and are not used to define the persona policy; they play no role in the counterfactual analysis. Instead, display names and birth dates do carry demographic information and serve as the signal for gender and age, respectively: their construction is described in Appendix~\ref{sec:signals}.

\xhdr{Treatment Conditions.}
In general, the choice of which demographic features to perturb is platform-dependent: the relevant signals are the ones that the platform can observe or infer.
On $\Xtwitter$, we consider $K = 3$ perturbations targeting location ($C_i = 1$), age ($C_i = 2$), and gender ($C_i = 3$). 
Each perturbation modifies exactly one demographic dimension while holding the behavior policy, political typology, and all remaining demographics at their baseline values; accounts with $C_i = 0$ serve as controls. 
Age is directly observed by the platform through the birth date declared at registration. Gender and location are not explicitly declared; instead, the platform can plausibly infer them from the
account's display name and IP address, respectively. 
While this cannot be verified under black-box access, demographic inference from names and IP addresses is standard practice in online advertising and content personalization. More details can be found in Appendix~\ref{sec:signals}. 

Within each persona $p$, we partition the $|\mathcal{I}_p| = 80$ replicas into 4 equal-sized groups of 20 accounts by balanced randomization, one group per condition $C \in \{0, 1, 2, 3\}$. With $P = 14$ personas, our experiment comprises $N = 1{,}120$ accounts in total. Crucially, each perturbation modifies exactly one demographic dimension, while holding the persona's political typology and all other demographics fixed at their baseline values. As the persona prompt is fixed by construction, observed differences in outcomes across conditions $C_i$ capture how the platform's content allocation responds to the perturbed attribute for a user following a fixed behavior policy~(Eq.~\eqref{eq:ate_persona}).

\subsection{Platform Interaction}
\label{sec:platform_interaction}
We now describe the interaction protocol for our deployment on {\Xtwitter}. While the experimental design above is mostly platform-agnostic, the choices below reflect the specific constraints of \Xtwitter.

\xhdr{Feed Interaction.}
We deploy the accounts from November 5 to November 28, 2024, spanning Election Day and its immediate aftermath. 
Before the experiment begins, each account follows a set of accounts selected during an onboarding phase: the LLM is presented with \Xtwitter's suggested-accounts page and chooses which accounts to follow given its persona.
Each account $i$ then conducts eight daily browsing sessions over the deployment period, with session start times sampled randomly throughout the day: four on the algorithmically ranked \emph{For You} feed and four on the reverse-chronological \emph{Following} feed. The feed is fixed for the duration of each session, which lasts approximately ten minutes.

In each session, the agent navigates to the platform's feed and observes content items $z_t$ served to it. %
For each item, the agent extracts the tweet's content and metadata from the rendered page and passes them to GPT-4o in a single API call together with the behavior policy prompt.
The model returns an action $b_t \in \{\text{like, follow, read-replies, ignore}\}$. The action is executed via browser automation, and the tuple $(z_t, b_t)$ is logged before advancing to the next item. A prompt-level instruction nudges the agent to take at least one active action every five items to ensure sufficient signal for the recommendation algorithm. Full feed engagement prompts and action definitions can be found in Appendix~\ref{sec:feed_engagement}. 

Once deployed, agents risk suspension if detected by the platform's systems. To mitigate detection, we introduce inter-action delays that simulate human browsing cadence, route each account through an IP address matching its assigned location, stagger account creation over time, and randomize session start times. Accounts may nonetheless face temporary locks requiring email verification, interrupting scheduled sessions and reducing observed exposures; we automate account recovery through a custom email-verification pipeline. Attrition rates are reported in Appendix~\ref{sec:attrition}; we find no evidence of differential attrition across treatment conditions. 

\xhdr{Data collection.}
We access the platform via Selenium, a framework to automate web browsers. %
This allows our sock puppets to observe the rendered feed as it would appear to an ordinary user.
For each content item $z_t$ served to account $i$, we record the item's text and author handle, its timestamp and engagement counts (likes, retweets, replies) at the time of observation, indicators for attached media, and the feed type (algorithmic or chronological). We also log the account's action $b_t \in \mathcal{B}$ and the full LLM output including its reasoning trace. These logs define the realized trajectories used to construct our metrics of interest (Section~\ref{sec:metrics}).

We define a content item as \emph{exposed} to an account if it is rendered during browsing and recorded by our logging pipeline. We do not impose a dwell-time threshold to define exposure: even when the agent chooses to ignore a piece of content, the LLM must process it to select the ignore action, so dwell time is mechanically driven by inference rather than serving as a measure of attention.

\section{Measurement and Estimation}
\label{sec:estimation}
This section describes our measurement and estimation framework. We begin by defining the outcome metrics used to characterize content exposure (Section~\ref{sec:metrics}). We then introduce the amplification ratio to compare content exposure on the algorithmic feed and chronological feed (Section~\ref{sec:amplification}). Finally, we present our counterfactual estimation strategy, first pooled across personas to capture average effects, then at the persona level to capture heterogeneity (Section~\ref{sec:counterfactual}).

\subsection{Metrics of Interest}
\label{sec:metrics}
We study how the platform’s content delivery varies across demographic profiles.
To this end, we focus on a set of metrics that have been widely used in prior audits of social media platforms. Note that our metrics are not specific to $\Xtwitter$. That said, we instantiate our discussion to the $\Xtwitter$ case study for clarity of presentation.

Classifying tweet content is a non-trivial task. For toxicity, reliable off-the-shelf classifiers exist and we adopt one directly. For political content and polarization, however, off-the-shelf models trained on general corpora performed poorly on our election-period data, likely due to the lack of domain-specific context. We therefore developed a custom LLM-based classification pipeline using chain-of-thought prompting \citep{wei2022chain}, adopting best practices for LLM-based annotation: explicit definitions, structured output constraints, and grounded few-shot examples \citep{baumann2025largelanguagemodelhacking}. Recent work has demonstrated that LLMs can match or exceed human annotators for this class of tasks
\citep{tornberg2023chatgpt, heseltine2024large}.
We validated our pipeline against a hand-labeled set of 200 tweets; full implementation details and prompts are reported in Appendix~\ref{sec:labels}.

\xhdr{Toxic Content.}
Toxicity has been a central outcome metric in prior algorithmic audits and content moderation studies
\citep{davidson2017automated, talat2016hateful, bouchaud2023crowdsourced}. Multiple off-the-shelf classifiers are available,
typically trained on large-scale annotated datasets of abusive or harmful language. We use the open-source \texttt{Detoxify} library \citep{Detoxify}, which outputs a
continuous toxicity score for each tweet. We define the binary outcome \emph{toxic} using a $0.5$ threshold.

\xhdr{Political Content.}
A tweet is labeled as \emph{political} if it clearly refers to: (i) the U.S.\ presidential election (e.g., campaigning, voting, results, legitimacy); (ii) U.S.\ political institutions, public policy, or governance; (iii) formal political actors such as candidates, elected officials, or political parties; or (iv) explicit partisan or electoral positioning. The classifier assigns a category from: $\{$\texttt{not\_political}, \texttt{political\_left}, \texttt{political\_neutral}, \texttt{political\_right}$\}$. From these we derive three binary outcomes: \emph{politicalness} (any political category, including neutral), \emph{right-leaning} (\texttt{political\_right} only), and \emph{left-leaning} (\texttt{political\_left} only), each computed over all tweets.
Our method for classifying political content, including the exact prompt specifications, can be found in Appendix~\ref{sec:labels}. 

\xhdr{Polarizing Content.}
A tweet is labeled as \emph{polarizing} if it satisfies two conditions: (1) it distinguishes between two or more social, political, religious, sexual, or racial groups (explicitly or implicitly), following the conceptual framework of \citet{naseem2025polarbenchmarkmultilingualmulticultural}, and (2) it frames those groups in a moralized conflict---such as blame, hostility, moral superiority, threat, or victim-oppressor narratives. Mentioning politics, religion, or identity alone is not sufficient; the tweet must actively pit groups against each other \citep{brady2021social, simchon2022troll}. When classified as polarizing, the model assigns a type from $\{$\texttt{political}, \texttt{religious}, \texttt{sexual}, \texttt{racial}, \texttt{other}$\}$. The categorical breakdown aids labeling accuracy by anchoring the model's reasoning; our analysis uses only the binary indicator \emph{polarizing} versus \emph{not polarizing}, pooling across all types. 
Our method for classifying polarizing content, including the exact prompt specifications, can be found in Appendix~\ref{sec:labels}. 

\xhdr{Author Reach.}
To capture whether the algorithmic feed favors content from high-reach accounts, we measure the follower count of each tweet's author as a proxy for author virality \citep{Ye2025}. We classify a tweet as \emph{high-reach} if its author's follower count exceeds the sample median, yielding a binary indicator analogous to the other outcome variables.

\subsection{``For You'' vs ``Following'' Feed Analysis}
\label{sec:amplification}
We measure whether $\Xtwitter$'s algorithmic ranking in the \emph{For You} feed amplifies exposure to specific content types relative to the reverse-chronological \emph{Following} feed. This comparison addresses long-standing concerns about filter bubbles and echo chambers~\citep{sunstein2018republic,rowland2011filter}: if algorithmic curation systematically over-represents certain content---political material, toxic language, or ideologically skewed viewpoints---the aggregate effect on public discourse may be substantial. %
These concerns have prompted a growing body of research ~\citep{Huszr2022,guess2023social,milli2025engagement} and press ~\citep{washpost2023foryou, wsj2024xalgorithm} to study whether and how algorithmic curation shapes content exposure.

\xhdr{Amplification Ratio.}
\citet{Huszr2022} introduced the notion of \emph{amplification ratio}: the relative likelihood that a set of tweets is seen under an algorithmically ranked feed compared to the reverse-chronological one. In their setting, the ratio is estimated via a randomized experiment where users are assigned exclusively to one of the two feeds. Our experimental design is different: each account browses both feeds during the deployment window, with sessions split evenly between the \emph{For You} and \emph{Following} feeds (Section~\ref{sec:methodology}). This yields \emph{paired observations} at the user level, enabling within-user comparisons that eliminate confounds from account history or unobserved user characteristics that might correlate with feed assignment in between-subjects designs.

Fix a content property of interest and let $\mathcal{Z}_0 \subset \mathcal{Z}$ denote the subset of content items satisfying that property (e.g., tweets classified as political). For account $i$ on day $d$, let $\textsc{fy}_{id}$ and $\textsc{fol}_{id}$ denote the sets of content items observed in the \emph{For You} and \emph{Following} feeds, respectively, with cardinalities $m_{id}^{\textsc{fy}} = |\textsc{fy}_{id}|$ and $m_{id}^{\textsc{fol}} = |\textsc{fol}_{id}|$.
Define the \emph{$\mathcal{Z}_0$-content rate} %
for account $i$ on day $d$ in each feed as:
    \begin{equation}\label{eq:content-rate}
        r_{id}^{\textsc{fy}} = \frac{1}{m_{id}^{\textsc{fy}}} \sum_{z \in \textsc{fy}_{id}} \mathbf{1}[z \in \mathcal{Z}_0], \qquad
        r_{id}^{\textsc{fol}} = \frac{1}{m_{id}^{\textsc{fol}}} \sum_{z \in \textsc{fol}_{id}} \mathbf{1}[z \in \mathcal{Z}_0].
    \end{equation}

We aggregate to the account level by averaging daily rates within each account, then compute the mean across accounts to obtain $\bar{r}^{\textsc{fy}}$ and $\bar{r}^{\textsc{fol}}$. The \emph{amplification ratio} for content property $\mathcal{Z}_0$ is:

\begin{equation}\label{eq:amplification-ratio}
      A(\mathcal{Z}_0) = \left( \frac{\bar{r}^{\textsc{fy}}}{\bar{r}^{\textsc{fol}}} - 1 \right) \times 100\%,
  \end{equation}
 
This approach first averages content rates within each account and then across accounts, ensuring equal weight per account regardless of the number of active days.\footnote{Because $\bar{r}^{\textsc{fol}}$ aggregates first over days and then over accounts, the denominator is well-defined whenever the content type has a nonzero base rate in the Following feed, a condition satisfied by all metrics we study.}
A value of $0\%$ indicates equal rates across feeds, while positive (negative) values indicate that the algorithmic feed \emph{amplifies} (\emph{de-amplifies}) the content type relative to the reverse-chronological baseline. Note that the two feeds differ not only in ranking but also in content pool, as the \emph{For You} feed surfaces content from outside the user's follow graph. Hence, our ratio captures the full change in exposure from switching feeds rather than isolating the ranking effect. %
Since each account browses both feeds, cross-feed spillovers may arise (e.g., engagement on one feed updating the platform's preference model). If present, such spillovers would likely pull the two feeds closer together, making our results conservative. We discuss robustness in more detail in Appendix~\ref{sec:amplification_more}.

We perform inference via persona-level cluster bootstrap: each draw resamples personas with replacement, preserving the within-persona dependence structure induced by shared behavioral policies.

\xhdr{Stratification by Ideology.}
A central question in the accountability literature is whether recommendation algorithms amplify \emph{congenial} content, i.e., content that aligns with users' political views. A growing body of evidence suggests they do, with a focus on political and polarizing content \citep{bakshy2015exposure, gonzalez2023asymmetric, bail2018exposure}. These findings motivate a stratified analysis: does amplification differ systematically for users with different ideological orientations?

Our experimental design is uniquely suited to answer this question. Traditional amplification studies face a fundamental identification challenge when stratifying by user ideology: the behaviors that reveal ideology to researchers (e.g., engagement patterns) are the same signals that platforms use to personalize content, creating circularity between the stratification variable and the outcome. Our design resolves this through an \emph{a priori} ideology definition: each persona is assigned a political orientation before any platform interaction occurs, and the LLM behavioral policy remains unchanged across all sessions. This clean separation enables a direct test of whether amplification is heterogeneous across user types.

We stratify \emph{left-leaning} and \emph{right-leaning} following the Pew typology (see Appendix~\ref{sec:pol}). We report the estimated amplification ratios stratified by ideology in Section~\ref{sec:fy-vs-following}. As in the aggregate analysis, we perform inference via persona-level cluster bootstrap. %

\subsection{Counterfactual Analysis}
\label{sec:counterfactual}

Having characterized the platform's relative amplification bias, we now turn to estimating the causal effect of user attributes on content exposure. We first estimate \emph{pooled} treatment effects to capture the average algorithmic response to demographic changes across all personas.
In Section~\ref{sec:setting}, we defined the outcome as a summary of the user's full trajectory; in practice, we estimate treatment effects at the account-day level. We aggregate binary tweet labels into daily exposure rates and define the \emph{algorithmic lift} as the additive difference between the exposure rate in the \emph{For You} feed and the \emph{Following} feed:
  \begin{equation}
  \label{eq:algorithmic-lift}
  y_{id} \;=\; r_{id}^{\textsc{fy}} \;-\; r_{id}^{\textsc{fol}},
  \end{equation}
  
where $r_{id}^{\textsc{fy}}$ and $r_{id}^{\textsc{fol}}$ are the content rates defined in Equation~\eqref{eq:content-rate}. The algorithmic lift $y_{id}$ is the dependent variable in all counterfactual regressions that follow. Intuitively, $y_{id}$ captures how much extra content of a type the algorithm injects beyond the chronological baseline \citep{milli2025engagement}. 
  
We choose this additive specification over the ratio used in Section~\ref{sec:amplification} for two reasons specific to the account-day unit of analysis. The first reason is {robustness to sparsity}: unlike the aggregate means, daily organic exposure to sparse content (e.g., toxic tweets) is frequently zero, making granular ratios undefined. The second reason is to focus on absolute differences: while the amplification ratio describes the system's \emph{relative bias}, the additive lift captures the magnitude of exposure. This allows us to quantify the actual ``dose'' of additional content of a certain type the algorithm injects into the user's experience, given a counterfactual perturbation.

\xhdr{Pooled Treatment Effects.}
To estimate the average effect of demographic signals, we encode the treatment \emph{direction} explicitly across personas:
\begin{itemize}
    \item \emph{Location}: $D^{\text{urb}}_i \in \{-1, 0, +1\}$, where $+1$ indicates the displayed state has a \emph{higher} urbanization rate than the persona's baseline state, $-1$ indicates a lower rate (more rural), and $0$ indicates baseline. We use the percentage of population residing in urbanized areas from the 2020 U.S.\ Census \citep{uscensus2020urban}.
    \item \emph{Age}: $D^{\text{age}}_i \in \{-1, 0, +1\}$, where $+1$ indicates the displayed age bracket exceeds the persona's baseline, $-1$ indicates it falls below, and $0$ indicates baseline.
    \item \emph{Gender}: $D^{\text{male}}_i = 1$ if account $i$ displays a male profile, $0$ if female.
\end{itemize}
Note that our ordinal encoding of age and location assumes a monotone relationship between these signals and algorithmic recommendations. Alternative specifications, such as dummies for specific categories, could capture non-monotone effects. 

Let $\mathbf{T}_i = (D^{\text{urb}}_i, D^{\text{age}}_i, D^{\text{male}}_i)^\top$ denote the treatment vector. We estimate the following model at the account-day level:
\begin{equation}
y_{id} \;=\; \gamma_d + \delta_{p_i} + \boldsymbol{\beta}^\top \mathbf{T}_i + X^\top_{id}\theta + \varepsilon_{id},
\label{eq:pooled_reg}
\end{equation}
where $\gamma_d$ are \emph{day} fixed effects, $\delta_{p_i}$ are \emph{persona} fixed effects, and $X_{id}$ is a vector of time-varying controls. The coefficient vector $\boldsymbol{\beta}$ identifies the average treatment effect of each demographic attribute on algorithmic lift, pooled across all personas. The covariate vector $X_{id}$ includes two sets of controls to improve precision: (1) \emph{Hour-of-day indicators} (i.e., binary indicators for session activity in each hour), and (2) \emph{Account age} (i.e., hours elapsed since account creation).%

Because treatment is randomly assigned within each persona,
$\boldsymbol{\beta}$ identifies the causal effect of each demographic attribute on algorithmic lift. We cluster standard errors at the account level to adjust for the correlation among repeated daily observations within the same account.

\xhdr{Heterogeneous Treatment Effects.}
The pooled analysis assumes that the platform's response to a demographic signal (e.g., an increase in age) is constant across all user profiles. However, algorithmic personalization is likely context-dependent. In this section, we relax the homogeneity assumption to estimate \emph{persona-specific} treatment effects.

We interact the treatment assignment with the persona indicator. Recall that $C_i \in \{0,1,\dots,K\}$ denotes the profile condition assigned to account $i$. Let $W_{i,p,k} = \mathbf{1}\{p_i = p, C_i = k\}$ indicate that account $i$ belongs to persona $p$ and was assigned to counterfactual condition $k$. We estimate:

\begin{equation}
y_{id}
\;=\;
\gamma_d
+ \delta_{p_i}
+ \sum_{p=1}^{P}\sum_{k=1}^{K}\beta_{p,k}\, W_{i,p,k}
+ X_{id}^\top\theta
+ \varepsilon_{id}.
\label{eq:het_reg}
\end{equation}

Here, $\beta_{p,k}$ captures the algorithmic lift for persona $p$ in condition $k$ relative to that persona's baseline. Under balanced randomization of $C_i$ within each persona, these coefficients identify the persona-specific average treatment effects on algorithmic lift. The control vector $X_{id}$ is identical to the one defined for the pooled analysis. We cluster standard errors at the account level.

\section{Case Study: Results from Deployment on \Xtwitter}
\label{sec:results}
We deployed $1,120$ synthetic accounts on \Xtwitter during the 2024 U.S. presidential election, as per the experimental design described in Section~\ref{sec:methodology}. This
section reports and discusses the main results, following the measurement and estimation framework of Section~\ref{sec:estimation}. 
First, we analyze whether \Xtwitter's algorithmic \emph{For You} feed
systematically amplifies specific content types relative to the chronological \emph{Following} feed (Section~\ref{sec:fy-vs-following}). Second, we estimate how perturbations to platform-visible demographic signals affect algorithmic content exposure (Section~\ref{sec:counterfactual_results}). %

\subsection{``For You'' vs ``Following'' Feed}
\label{sec:fy-vs-following}

We first assess the baseline behavior of the recommendation algorithm by computing the \emph{amplification ratio} for different content categories. This within-subject comparison controls for user-level heterogeneity by comparing the algorithmic feed against the user's own chronological baseline. Results are reported in Table~\ref{tab:amplification-results}.

  \begin{table}[t]
      \centering
      \small
      \begin{tabular}{lcccc}
      \toprule
      \textbf{Content} & \textbf{Baseline ($\bar{r}^{\textsc{FOL}}$)} & \textbf{Amplification} &
  \textbf{CI (low)} & \textbf{CI (high)} \\
      \midrule
      Political & $38.0\%$ & $+14.2\%$ & $+9.1\%$ & $+19.8\%$ \\
      Right-leaning & $14.9\%$ & $+23.3\%$ & $+15.8\%$ & $+31.9\%$ \\
      Left-leaning & $6.5\%$ & $-6.1\%$ & $-14.6\%$ & $+2.1\%$ \\
      Toxic & $3.9\%$ & $+39.2\%$ & $+14.4\%$ & $+69.7\%$ \\
      Polarizing & $14.0\%$ & $+32.3\%$ & $+23.7\%$ & $+41.6\%$ \\
      High-reach & $64.5\%$ & $+3.4\%$ & $-6.1\%$ & $+16.7\%$ \\
      \bottomrule
      \end{tabular}
      \begin{minipage}{\columnwidth}
      \vspace{0.5em}
      \caption{Amplification of content types in X's algorithmic \emph{For You} feed relative to the \emph{Following} feed. The baseline column reports the average content rate in the \emph{Following} feed ($\bar{r}^{\text{FOL}}$). Amplification values represent the relative percentage difference in content exposure. \textbf{The \emph{For You} feed exposes users to significantly more toxic, polarizing, political, and right-leaning content.} Confidence intervals (95\%) are constructed via persona-level cluster bootstrap.}
      \label{tab:amplification-results}
      \end{minipage}
    \end{table}

We find that all content types except left-leaning and high-reach are significantly amplified in the For You feed. The largest effects are for toxic (+39.2\%) and polarizing (+32.3\%) content, consistent
with evidence that algorithmic ranking increases exposure to anger and out-group hostility~\citep{milli2025engagement}. 
Political content is amplified by +14.2\%; its high baseline prevalence makes this the largest shift in absolute terms. 
We also observe a statistically significant ideological asymmetry in the aggregate results: right-leaning content is amplified, whereas left-leaning content shows no significant amplification ($+23.3\%$ vs $-6.1\%$, with the 95\% CI including zero for the latter).  As we demonstrate next, this aggregate pattern masks sharper divergences across user groups.

\begin{figure}[htbp]
  \centering
  \includegraphics[width=0.9\textwidth]{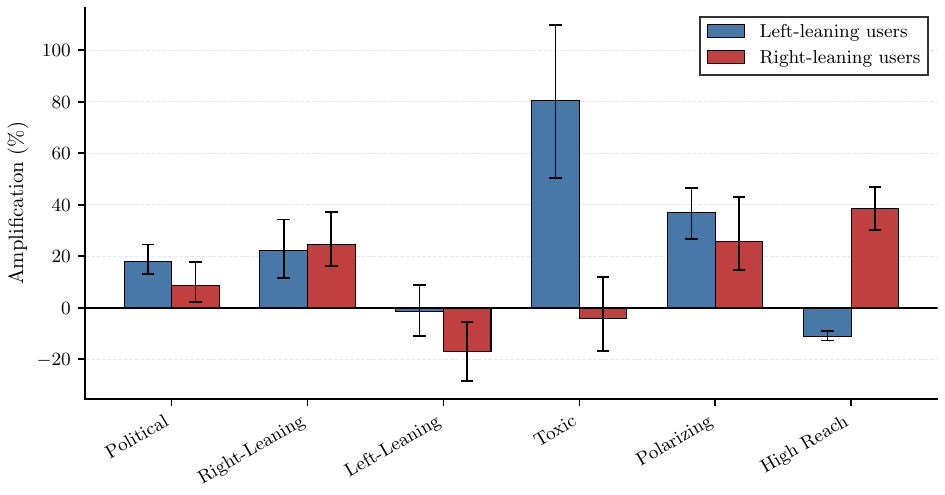}
\caption{Amplification ratios by content type, stratified by user political leaning (left: blue, right: red). Bars show point estimates. \textbf{Toxic content is amplified by 80\% for left-leaning users but not for right-leaning users;
right-leaning content is amplified for both groups, while left-leaning content is suppressed for right-leaning users.} Error bars indicate 95\% confidence intervals. }
  \label{fig:amplification}
\end{figure}

\xhdr{Stratification by Ideology.}
Table~\ref{tab:amplification-leaning} and Figure~\ref{fig:amplification} stratify these results by the political leaning of the account. The aggregate patterns mask substantial heterogeneity: amplification effects differ sharply---and for some content types, even reverse sign---across
left-leaning and right-leaning users.

The most striking example is for toxic content: the algorithm amplifies it by +80.3\% for left-leaning users but shows no significant effect for right-leaning users ($-4.0\%$, CI includes zero)---a gap of $84.3$ percentage points ($p < 0.01$). A similar pattern holds for political content, which is amplified roughly twice as much for left-leaning users as for right-leaning users ($+18.1\%$ vs.\ $+8.7\%$), though the difference is only marginally significant. Polarizing content is also amplified more for left-leaning users ($+36.9\%$ vs.\ $+25.8\%$), though the difference is not statistically significant. High-reach content shows the opposite pattern, with amplification concentrated among right-leaning users ($+38.4\%$ vs.\ $-11.0\%$, $p < 0.01$).

Right-leaning content is amplified for \emph{both} user groups ($+22.3\%$ for left-leaning users, $+24.5\%$ for right-leaning users), with no significant difference between them. Left-leaning content, by contrast, shows no amplification even for left-leaning users ($-1.4\%$, not significant) and is actively suppressed for right-leaning users ($-17.0\%$). The uniform amplification of right-leaning content across user types is consistent with a level effect in the algorithm's content allocation, not with personalization toward ideologically congruent material. For right-leaning users, this effect is reinforced by the suppression of left-leaning content, creating an asymmetric echo chamber that does not arise for left-leaning users. This result reveals striking disparate exposure: to put it plainly, the algorithmic feed amplifies right-leaning content for left-leaning users by more than $22\%$, while it suppresses left-leaning content by $17\%$ for right-leaning users.

Across the hypothesis tests in Tables~\ref{tab:amplification-results} and~\ref{tab:amplification-leaning}, all significant results survive Benjamini--Hochberg (BH) correction for multiple hypothesis testing \citep{benjamini1995controlling}, with the exception of the political between-group difference.

  \begin{table}[t]
  \centering
  \small
  \begin{tabular}{lccc ccc l}
  \toprule
  & \multicolumn{3}{c}{\textbf{Left-leaning users}}
  & \multicolumn{3}{c}{\textbf{Right-leaning users}}
  & \textbf{Difference} \\
  \cmidrule(lr){2-4} \cmidrule(lr){5-7}
  \textbf{Content}
  & \textbf{Ampl.} & \textbf{CI (low)} & \textbf{CI (high)}
  & \textbf{Ampl.} & \textbf{CI (low)} & \textbf{CI (high)}
  & \\
  \midrule
  Political
    & $+18.1\%$ & $+13.1\%$ & $+24.5\%$
    & $+8.7\%$  & $+2.1\%$  & $+17.8\%$
    & $-9.4$ pp$^{*}$ \\
  Right-leaning
    & $+22.3\%$ & $+11.5\%$ & $+34.2\%$
    & $+24.5\%$ & $+16.2\%$ & $+37.2\%$
    & $+2.2$ pp \\
  Left-leaning
    & $-1.4\%$  & $-10.9\%$  & $+8.9\%$
    & $-17.0\%$ & $-28.5\%$ & $-5.5\%$
    & $-15.6$ pp$^{**}$ \\
  Toxic
    & $+80.3\%$ & $+50.4\%$ & $+109.7\%$
    & $-4.0\%$  & $-16.7\%$ & $+12.0\%$
    & $-84.3$ pp$^{***}$ \\
  Polarizing
    & $+36.9\%$ & $+26.6\%$ & $+46.6\%$
    & $+25.8\%$ & $+14.7\%$ & $+43.1\%$
    & $-11.0$ pp$^{}$ \\
  High-reach
    & $-11.0\%$  & $-12.8\%$  & $-9.0\%$
    & $+38.4\%$ & $+30.3\%$ & $+46.8\%$
    & $+49.4$ pp$^{***}$ \\
  \bottomrule
  \end{tabular}
  \begin{minipage}{\columnwidth}
  \vspace{0.5em}
    \caption{Amplification ratios stratified by political leaning of users. \textbf{The algorithm's effects differ substantially across user groups.} The difference column reports the gap (Right $-$ Left) in percentage points (pp). Confidence intervals (95\%) are constructed via persona-level cluster bootstrap; p-values: $^{*}p<0.10$, $^{**}p<0.05$, $^{***}p<0.01$.}
  \label{tab:amplification-leaning}
  \end{minipage}
\end{table}

\subsection{Counterfactual Analysis}
\label{sec:counterfactual_results}
We now turn to the counterfactual question: does the algorithm respond differently to users with different demographic profiles, \emph{holding behavior fixed}? Using the algorithmic lift $y_{id}$ defined in Eq.~\eqref{eq:algorithmic-lift} as the dependent variable, we estimate the causal effect of perturbing age, gender, and location signals on the gap between algorithmic and chronological content exposure.

\xhdr{Pooled Treatment Effects.}
\label{sec:pooled-results}
Table~\ref{tab:pooled-ate} reports pooled treatment effects from the model in Equation~\eqref{eq:pooled_reg}. The main finding is that demographic perturbations have limited systematic effects on algorithmic lift when pooled across user profiles: out of the estimated coefficients, only two are statistically significant at the 5\% level. In particular, we find that displaying an older age increases the algorithmic lift of toxic content by $0.80$ percentage points ($p = 0.031$), and displaying a more urban location reduces the lift of polarizing content by $1.60$ percentage points ($p = 0.024$). Relative to the baseline algorithmic lift, these approximately correspond to a $70\%$ increase in the algorithm's additional injection of toxic content (baseline lift: 1.15 pp) and a $38\%$ reduction for polarizing content (baseline lift: 4.18 pp). Gender has no significant effect on any outcome. However, neither result survives BH correction for multiple testing across the 18 hypotheses.

\begin{table}[t]
    \centering
    \small
    \begin{tabular}{l cc cc cc}
    \toprule
    & \multicolumn{2}{c}{\textbf{Location} ($D^{\text{urb}}$)}
    & \multicolumn{2}{c}{\textbf{Age} ($D^{\text{age}}$)}
    & \multicolumn{2}{c}{\textbf{Gender} ($D^{\text{male}}$)} \\
    \cmidrule(lr){2-3} \cmidrule(lr){4-5} \cmidrule(lr){6-7}
    \textbf{Content}
    & $\hat{\beta}$ & \textbf{95\% CI}
    & $\hat{\beta}$ & \textbf{95\% CI}
    & $\hat{\beta}$ & \textbf{95\% CI} \\
    \midrule
    Political
      & $-1.70$ & {\scriptsize$[-3.86,\,0.46]$}
      & $0.70$  & {\scriptsize$[-1.26,\,2.66]$}
      & $0.90$  & {\scriptsize$[-1.45,\,3.25]$} \\
      & {\scriptsize$(0.142)$} &
      & {\scriptsize$(0.473)$} &
      & {\scriptsize$(0.449)$} & \\[2pt]

    Right-Leaning
      & $-0.50$ & {\scriptsize$[-1.87,\,0.87]$}
      & $1.00$  & {\scriptsize$[-0.18,\,2.18]$}
      & $-0.40$ & {\scriptsize$[-1.58,\,0.78]$} \\
      & {\scriptsize$(0.477)$} &
      & {\scriptsize$(0.102)$} &
      & {\scriptsize$(0.534)$} & \\[2pt]

    Left-Leaning
      & $-0.70$ & {\scriptsize$[-1.48,\,0.08]$}
      & $-0.30$ & {\scriptsize$[-1.08,\,0.48]$}
      & $0.30$  & {\scriptsize$[-0.48,\,1.08]$} \\
      & {\scriptsize$(0.131)$} &
      & {\scriptsize$(0.367)$} &
      & {\scriptsize$(0.436)$} & \\[2pt]

    Toxic
      & $0.10$           & {\scriptsize$[-0.68,\,0.88]$}
      & $\mathbf{0.80}$  & {\scriptsize$\mathbf{[0.02,\,1.58]}$}
      & $-0.10$          & {\scriptsize$[-0.69,\,0.49]$} \\
      & {\scriptsize$(0.770)$} &
      & {\scriptsize$\mathbf{(0.031)}$} &
      & {\scriptsize$(0.876)$} & \\[2pt]

    Polarizing
      & $\mathbf{-1.60}$ & {\scriptsize$\mathbf{[-2.97,\,-0.23]}$}
      & $0.50$           & {\scriptsize$[-0.68,\,1.68]$}
      & $-0.70$          & {\scriptsize$[-2.07,\,0.67]$} \\
      & {\scriptsize$\mathbf{(0.024)}$} &
      & {\scriptsize$(0.439)$} &
      & {\scriptsize$(0.302)$} & \\[2pt]

    High-reach
      & $1.30$  & {\scriptsize$[-1.25,\,3.85]$}
      & $-0.90$ & {\scriptsize$[-3.64,\,1.84]$}
      & $-0.10$ & {\scriptsize$[-2.84,\,2.64]$} \\
      & {\scriptsize$(0.334)$} &
      & {\scriptsize$(0.524)$} &
      & {\scriptsize$(0.939)$} & \\
    \bottomrule
    \end{tabular}
  \begin{minipage}{\columnwidth}
  \vspace{0.5em}
    \caption{Pooled treatment effects on algorithmic lift. Each
   coefficient represents the average effect of a one-unit change in the demographic signal on the gap
  between \emph{For You} and \emph{Following} content rates. Coefficients for binary outcomes are scaled
  by 100 and expressed in percentage points. \textbf{Most pooled effects are small and not statistically
  significant, suggesting the algorithm's response to demographic signals is not systematic across user
  profiles.} Values in parentheses are $p$-values
  computed from standard errors clustered at the account level. Significant coefficients ($p < 0.05$) shown in bold.}
  \label{tab:pooled-ate}
  \end{minipage}
\end{table}

The absence of strong pooled effects does not imply that the platform ignores demographic signals entirely. Rather, it is consistent with the hypothesis that any algorithmic responsiveness to profile attributes is \emph{persona-dependent}, i.e., conditioned on the full behavioral profile the platform has accumulated. We investigate this possibility next.

\xhdr{Heterogeneous Treatment Effects.}
\label{sec:heterogeneous-results}
The near-zero pooled estimates could reflect genuinely null effects or heterogeneity that cancels in the aggregate. To distinguish these, we estimate persona-specific treatment effects via the model of Equation~\eqref{eq:het_reg}. Joint $F$-tests reject the null of zero treatment effects for all outcomes at the 5\% level: left-leaning ($F = 2.02$, $p < 0.001$), politicalness ($F = 1.93$, $p < 0.001$), author reach ($F = 1.93$, $p < 0.001$), toxic ($F = 1.89$, $p < 0.001$), right-leaning ($F = 1.57$, $p = 0.011$), and polarizing ($F = 1.49$, $p = 0.022$). All six remain significant after BH correction. 

\begin{figure}[htbp]
  \centering
  \includegraphics[width=\textwidth]{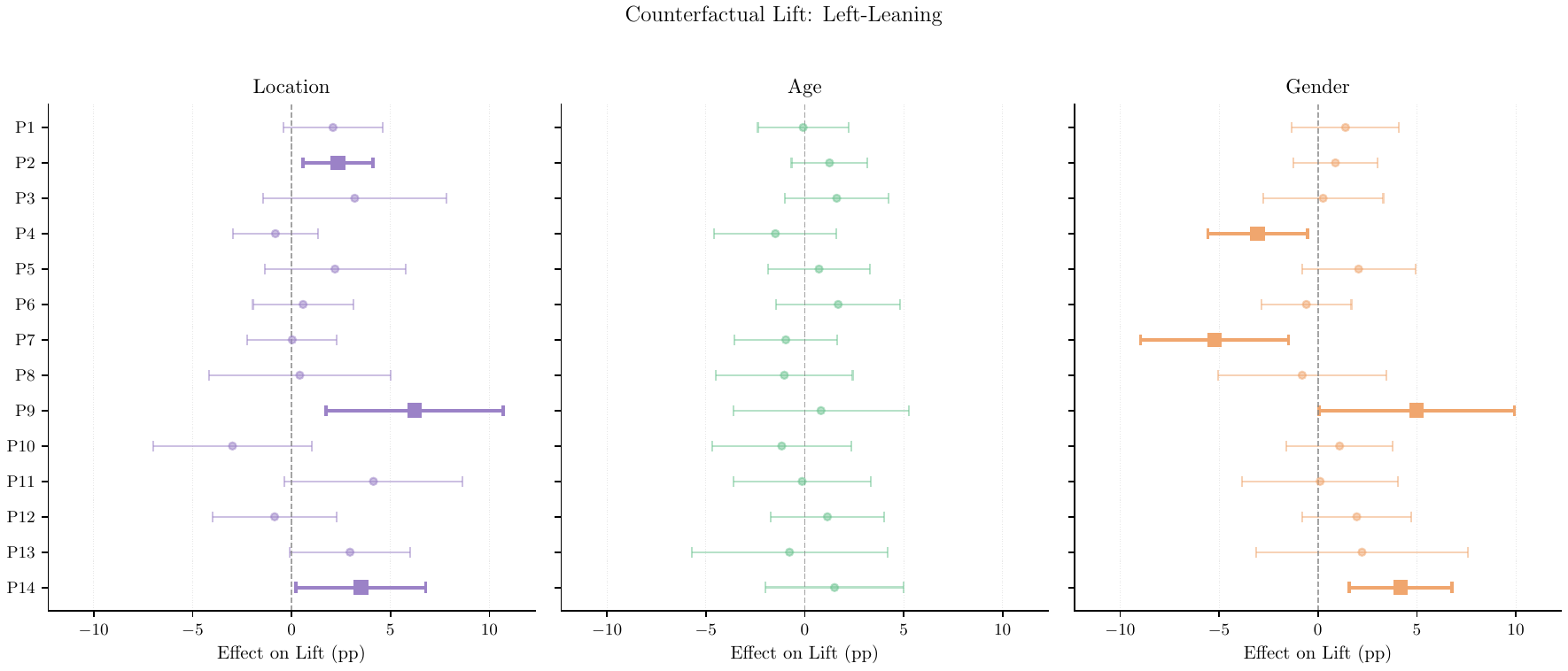}
  \caption{Persona-specific treatment effects on algorithmic lift for left-leaning content, estimated via model~\eqref{eq:het_reg}. Each row is a persona,  each panel a treatment. Points are estimated coefficients with 95\% confidence intervals (filled markers indicate significance). \textbf{Left-leaning content exhibits the most persona-level heterogeneity: all pooled effects are null, yet 7 persona-specific effects are significant.}}
  \label{fig:forest-left}
\end{figure}

Figure~\ref{fig:forest-left} illustrates this pattern for left-leaning content, the outcome with the strongest heterogeneity. None of the three pooled coefficients for left-leaning content are significant (Table~\ref{tab:pooled-ate}), yet persona-specific effects vary in sign and magnitude across personas, so that no consistent average effect emerges in the pooled estimate. Gender and location perturbations drive most of the heterogeneity, particularly for left-leaning and polarizing content; age effects are largely confined to toxicity and author reach. Full persona-level results are reported in Appendix~\ref{sec:heterogeneous-ate}.

At the individual level, 26 of 252 persona--outcome--treatment coefficients are significant at the 5\% level.
Given the number of comparisons, these individual coefficients should not be interpreted in isolation. Instead, together with the joint rejection of homogeneity and the near-zero pooled estimates, they indicate that the algorithm's response to demographic signals varies in direction across user types.
This has a direct methodological implication: pooled audits that average treatment effects across heterogeneous user populations may fail to detect algorithmic discrimination present for specific subgroups.

\subsection{Summary of Findings}
Taken together, the amplification analysis of Section~\ref{sec:fy-vs-following} and the counterfactual analysis of Section~\ref{sec:counterfactual_results} point to two main conclusions.
First, consuming content through {\Xtwitter}'s algorithmic \emph{For You} feed leads to substantially higher exposure to toxic, polarizing, and right-leaning content than relying on the chronological \emph{Following} feed.
This amplification gap is uneven across user groups: left-leaning users are disproportionately exposed to toxic content, while right-leaning users experience a one-sided information environment in which politically congenial content is amplified and opposing content is suppressed. Second, perturbing observable demographic signals causally affects algorithmic content delivery, but in a heterogeneous, persona-dependent manner. Pooled treatment effects are largely null, yet joint tests reject the hypothesis of homogeneous effects for all outcomes. The same perturbation can shift content delivery in opposite directions depending on the user's behavioral profile, producing effects that cancel in the aggregate.

A common pattern emerges across both analyses: population-level summaries mask sharp differences in how the algorithm treats distinct user groups. These findings highlight the value of moving beyond aggregate metrics in algorithmic auditing, toward designs that can detect heterogeneous effects across demographically and behaviorally distinct subpopulations.

\section{Discussion}
\label{sec:discussion}

In this paper, we introduced a framework that uses LLM-powered agents for black-box auditing of personalization systems. Our agents are instantiated to follow certain ``personas'', and reason about the content they are exposed to before engaging. Both observable and latent attributes can be encoded directly in the agent's prompt, enabling stratification by design rather than through post-hoc inference from behavioral proxies. Crucially, because behavior is fully specified by the prompt, it is decoupled from user attributes: platform-visible signals can then be perturbed across conditions without altering how the agent behaves, enabling counterfactual auditing. 

We deployed our auditing framework on {\Xtwitter} during the 2024 U.S.\ presidential election, operating 1{,}120 agents and collecting over 200{,}000 impressions. %
Consistent with prior work and anecdotal evidence, we find that the algorithmic feed amplifies toxic, polarizing, political, and right-leaning content relative to the chronological baseline; left-leaning content is not significantly amplified. Stratifying by ideology reveals highly asymmetric amplification: right-leaning content is amplified regardless of user leaning, while left-leaning content is suppressed for right-leaning users. 
In our counterfactual analysis, pooled demographic effects are largely null, but persona-level analyses
reject homogeneity of treatment effects for all outcomes, indicating that the algorithm's response to demographic signals is persona-dependent.

\xhdr{Limitations.}
Our study is subject to a few limitations. 
First, GPT-4o (which was the LLM that we used for our agents) lacks human lived experience and might carry biases when choosing how to react to certain content, given the persona it has been instructed to follow. To take this point one step further, we do not know the extent to which GPT-4o (with the prompt instructions that we gave it) is the best possible LLM to simulate human behavior in social media interactions. In fact, there has been a lot of recent, exciting work~\citep{wu2026humanlm} on identifying ways to make LLM agents better imitate humans making choices. All that being said, the validity and importance of our framework does not require LLMs to faithfully approximate humans: it requires a \emph{fixed} behavioral policy across replicas and behavior realistic enough to trigger personalization. For the counterfactual analysis, any additive LLM bias cancels under within-persona differencing; for the amplification analysis, the within-user comparison remains internally valid, but LLM engagement biases may shape the feedback loop and hence the levels of algorithmic amplification.

With regards to our deployment on $\Xtwitter$, given that we have 20 accounts per persona-condition cell and 6 outcome variables across three treatments, we test a large number of hypotheses relative to the available observations; we therefore present the heterogeneous analysis as exploratory, with evidence for heterogeneity resting on joint $F$-tests rather than individual coefficients. %

Finally, our findings are specific to
{\Xtwitter} during a period of exceptional political salience (recall, we ran our study very soon after the 2024 US presidential election), and all of our accounts are newly created. As a result, the algorithm may weight profile signals differently than for established users.

\xhdr{Future Work.}
Our work opens up several avenues for future research. Scaling to hundreds of personas would sharpen estimates of demographic effects; with sufficient power, this would also enable moderation analysis to identify which persona attributes (e.g., engagement patterns or political ideology) interact with the algorithm to shape its response to demographic signals. 
Enabling agents to \emph{create} content (posting tweets, replying, quote-tweeting) would generate richer engagement signals and make the feedback loop between agent and algorithm more realistic, likely triggering deeper personalization. This richer interaction mode would require substantially more careful IRB oversight, as agent-generated content could directly affect real users.

Perhaps the most important extension is deploying our auditing framework on other platforms. Nothing in our design is specific to {\Xtwitter}; the same approach applies to any platform with an algorithmic feed. The need for such audits is growing: policymakers increasingly demand algorithmic accountability, and platforms themselves have begun to respond. {\Xtwitter}, for instance, recently published the source code of its recommendation pipeline \citep{xlink}. However, code alone does not make a system transparent: without model weights, production data, and surrounding infrastructure, the released code cannot be independently verified.
Moreover, the patterns that audits seek to detect are often emergent consequences of optimization objectives instead of behaviors readable from source code. Black-box auditing therefore remains necessary to
characterize what users actually experience.

Finally, we stress that our framework need not be adversarial: platforms themselves could deploy synthetic agent populations in sandboxed environments to stress-test algorithm changes before live rollout, or regulators could run standardized agent cohorts to produce auditable, reproducible compliance evidence.

\section*{Acknowledgments}
This work was generously supported by a MacArthur Foundation grant and by an MIT MGAIC grant.

\bibliographystyle{abbrvnat}
\bibliography{references}

\appendix
\newpage
\begin{center}
  \Large\textbf{Supplementary Material}
\end{center}
\section{Details on Persona Creation}

\subsection{Full Profiles}
Table~\ref{tab:persona-cf} details the three counterfactual perturbations applied to each
persona (location, age, and gender) relative to the base profile. Table~\ref{tab:persona-invariant}
reports the attributes held fixed across all conditions within each persona (race/ethnicity, education, and political leaning).

 \begin{table}[htbp]
  \begin{tabular}{clll}
  \toprule
  Persona & Location (C=1) & Age (C=2) & Gender (C=3) \\
  \midrule
  1  & Arizona $\rightarrow$ Texas & 30--49 $\rightarrow$ 65+ & M $\rightarrow$ F \\
  2  & Florida $\rightarrow$ Colorado & 30--49 $\rightarrow$ 18--29 & M $\rightarrow$ F \\
  3  & Florida $\rightarrow$ North Carolina & 18--29 $\rightarrow$ 30--49 & F $\rightarrow$ M \\
  4  & Louisiana $\rightarrow$ New York & 30--49 $\rightarrow$ 50--64 & F $\rightarrow$ M \\
  5  & Massachusetts $\rightarrow$ California & 18--29 $\rightarrow$ 50--64 & F $\rightarrow$ M \\
  6  & Alabama $\rightarrow$ Illinois & 50--64 $\rightarrow$ 18--29 & M $\rightarrow$ F \\
  7  & Michigan $\rightarrow$ Washington & 18--29 $\rightarrow$ 30--49 & M $\rightarrow$ F \\
  8  & Nevada $\rightarrow$ Texas & 30--49 $\rightarrow$ 65+ & M $\rightarrow$ F \\
  9  & Tennessee $\rightarrow$ Arizona & 18--29 $\rightarrow$ 65+ & F $\rightarrow$ M \\
  10 & Indiana $\rightarrow$ California & 30--49 $\rightarrow$ 18--29 & M $\rightarrow$ F \\
  11 & California $\rightarrow$ Florida & 50--64 $\rightarrow$ 30--49 & M $\rightarrow$ F \\
  12 & California $\rightarrow$ California & 18--29 $\rightarrow$ 30--49 & M $\rightarrow$ F \\
  13 & New York $\rightarrow$ Ohio & 30--49 $\rightarrow$ 50--64 & F $\rightarrow$ M \\
  14 & California $\rightarrow$ Illinois & 30--49 $\rightarrow$ 18--29 & F $\rightarrow$ M \\
  \bottomrule
  \end{tabular}
  \vspace{0.2em}
\caption{Counterfactual perturbations by persona. Each counterfactual changes exactly one
platform-visible attribute relative to the base profile. Locations are reported at the state level (cities omitted), ages are sampled uniformly from the buckets.}
  \label{tab:persona-cf}
  \end{table}

  \begin{table}[htbp]
  \begin{tabular}{clll}
  \toprule
  Persona & Race/Ethnicity & Education & Political Leaning \\
  \midrule
  1  & White (Hispanic) & Associate's & Ambivalent Right \\
  2  & White (Non-Hisp.) & Some college & Ambivalent Right \\
  3  & American Indian & Some college & Establishment Liberals \\
  4  & White (Non-Hisp.) & High school & Democratic Mainstays \\
  5  & White (Non-Hisp.) & High school & Outsider Left \\
  6  & White (Non-Hisp.) & Some college & Progressive Left \\
  7  & White (Non-Hisp.) & 11th grade & Outsider Left \\
  8  & White (Hispanic) & High school & Faith and Flag Conservatives \\
  9  & Black & Bachelor's & Outsider Left \\
  10 & White (Non-Hisp.) & Some college & Populist Right \\
  11 & Black & Bachelor's & Outsider Left \\
  12 & White (Non-Hisp.) & 10th grade & Ambivalent Right \\
  13 & White (Non-Hisp.) & Bachelor's & Committed Conservatives \\
  14 & Asian & High school & Progressive Left \\
  \bottomrule
  \end{tabular}
  \vspace{0.2cm}
\caption{Unperturbed persona attributes. These attributes define the behavioral policy, together with the base values of the platform-visible attributes.}
\label{tab:persona-invariant}
  \end{table}

\subsection{Demographic Sampling Distributions}
\label{sec:demographics}

Each persona's demographic attributes are sampled independently from real-world population distributions. We describe the source and sampling weights for each attribute below. All the external information is either derived from the U.S.\ Census \citep{uscensus2021demographics} or Pew Research Center \citep{pew2021typology}.

\xhdr{Gender.}
Gender is drawn from U.S.\ Census population: Female (50.5\%), Male (47.2\%), Other (2.3\%).

\xhdr{Age.}
Age brackets are sampled by combining Census population shares with platform-specific usage rates. Let $P(\text{age})$ denote the Census share and $P(\text{Twitter} \mid \text{age})$ the platform usage rate. We compute $P(\text{age} \mid \text{Twitter}) \propto P(\text{Twitter} \mid \text{age}) \cdot P(\text{age})$, yielding the following sampling weights:

\begin{center}
\begin{tabular}{lcccc}
\toprule
& 18--29 & 30--49 & 50--64 & 65+ \\
\midrule
Census share $P(\text{age})$ & 13.3\% & 25.8\% & 19.0\% & 17.1\% \\
Usage rate $P(\text{Twitter} \mid \text{age})$ & 84\% & 81\% & 73\% & 45\% \\
Sampling weight $P(\text{age} \mid \text{Twitter})$ & 20.8\% & 38.9\% & 25.8\% & 14.4\% \\
\bottomrule
\end{tabular}
\end{center}

Within each bracket, a specific age is drawn uniformly at random.

\xhdr{Race and ethnicity.}
Race is drawn from Census proportions, collapsed to seven categories: White Non-Hispanic (58.9\%), White Hispanic (16.6\%), Black (13.6\%), Asian (6.3\%), American Indian (1.3\%), Native Hawaiian/Pacific Islander (0.3\%), and Multiracial (3.0\%).

\xhdr{Location.}
Location is drawn from U.S.\ incorporated places with population exceeding 50,000, with sampling probability proportional to city population (2022 Census estimates, 333 cities).

\xhdr{Education.}
Education is sampled conditional on the persona's age, gender, and race using Census educational attainment tables, covering 15 levels: none, 1st--4th grade, 5th--6th grade, 7th--8th grade, 9th grade, 10th grade, 11th grade, high school graduate, some college (no degree), occupational Associate's, academic Associate's, Bachelor's, Master's, professional degree, and doctoral degree.

\xhdr{Political typology.}
Political typology is sampled from $P(\theta \mid \text{age}, \text{gender})$, derived from the demographic breakdowns published in Pew's Political Typology report \citep{pew2021typology}. We provide more details on this in the next section.

\subsection{Political Typology Descriptions}
\label{sec:pol}
We adopt the nine-category political typology developed by Pew Research Center \citep{pew2021typology}, which classifies Americans based on their political values and attitudes rather than party affiliation alone. The following descriptions are drawn directly from Pew's
published characterizations.

\xhdr{Faith and Flag Conservative.}
Faith and Flag Conservatives are highly religious, politically engaged and both socially and economically conservative. They favor a robust role for religion in public life and a smaller role for government in society, and they hold that a strong American military is essential in international affairs. They overwhelmingly identify with the GOP and remain strong supporters of former President Donald Trump. More than four-in-ten are White evangelical Protestants, the highest share of any political typology group. Faith and Flag Conservatives are more likely than those in other groups to emphasize the importance of religion in their lives and to hold restrictive attitudes on abortion and same-sex marriage. Three-quarters say that the best way to ensure peace is through military strength rather than through diplomatic means. Faith and Flag Conservatives are the oldest typology group, with a median age of 57.

\xhdr{Committed Conservative.}
Staunchly conservative and overwhelmingly Republican, Committed Conservatives hold pro-business views traditionally associated with the Republican Party, have favorable attitudes about international trade and favor a limited role of government. Their approach to international relations centers on engaging with U.S.\ allies and maintaining American military might. Committed Conservatives tend to hold more moderate positions on immigration than the two other deeply conservative groups. While nearly all voted for Donald Trump for president in 2020, and most hold positive views of him today, Committed Conservatives are less likely than Populist Right and Faith and Flag Conservatives to favor a major role for Trump in their party's future. Committed Conservatives are among the most educated of the GOP-oriented groups and are among the highest-income political typology groups.

\xhdr{Populist Right.}
Very conservative and overwhelmingly Republican, Populist Right hold highly restrictive views about immigration policy and are very critical of government. But, in contrast to other parts of the GOP coalition, their criticism extends well beyond government to views of big business and to the economic system as a whole: 82\% say that large corporations are having a negative impact on the way things are going in the country, and nearly half support higher taxes on the wealthy and on large corporations. A majority of Populist Right are women (54\%). Populist Right are also one of the least highly educated groups; just two-in-ten are college graduates. Nearly nine-in-ten say that the economic system in this country unfairly favors powerful interests, far higher than the share in any other Republican-oriented group.

\xhdr{Ambivalent Right.}
On issues ranging from the size of the federal government to views about business, gender and race, Ambivalent Right hold many views that are largely consistent with core conservative values. Yet they also hold more moderate stances on several social issues and differ from some other segments of the GOP coalition in taking a more internationalist view of foreign policy and a less restrictive position on immigration. With 63\% under the age of 50, they are substantially younger than other Republican-oriented groups. About two-thirds (65\%) are White, making this group more racially and ethnically diverse than other GOP coalition groups. Just over half (54\%) say abortion should be legal in all or most cases, a view held by far smaller shares of those in other GOP groups. They are also distinct from other Republican-oriented groups in their views of Donald Trump---more likely to say they feel coldly toward the former president (46\%) than warmly (34\%).

\xhdr{Stressed Sideliner.}
Stressed Sideliners are generally disconnected from politics and the two major parties, voting at lower rates than most other typology groups. Although Stressed Sideliners make up 15\% of American adults, they were just 10\% of 2020 voters due to their relatively low turnout rate. They are split evenly between those who identify with or lean toward the Republican Party (45\%) and those who are Democrats or Democratic leaners (45\%). They tend to lean liberal on economic issues and tilt conservative on some social issues. About four-in-ten (43\%) live in lower-income households, higher than most other political typology groups. They are the group most likely to describe their personal financial situation as only fair or poor (63\%). Stressed Sideliners are also one of the least highly educated groups.

\xhdr{Outsider Left.}
Outsider Left are the youngest typology group, making up 10\% of the public. Holding liberal views on most issues and overwhelmingly voting Democratic, they aren't particularly enamored with the Democratic Party---though they have deeply negative views of the GOP. Nearly half (48\%) describe their own political views as liberal. Outsider Left are somewhat more skeptical of government's role than other Democratic-oriented groups. Four-in-ten are under the age of 30 and 83\% are under 50. They are racially and ethnically diverse: about half (49\%) are White, 20\% are Hispanic, 15\% are Black and 10\% are Asian. Just 53\% say the Democratic Party represents them at least somewhat well, and an overwhelming majority (86\%) say that they usually feel like none of the candidates for public office represent their views well.

\xhdr{Democratic Mainstay.}
Democratic Mainstays are one of the largest groups in the political typology and the largest single group as a share of the Democratic coalition. They generally favor policies that expand the social safety net and support higher taxes on corporations. But they are somewhat more hawkish than other Democratic-oriented groups on foreign policy and less liberal on immigration policy and some social issues. Nearly half (49\%) consider themselves ``strong Democrats.'' Democratic Mainstays are slightly older and have less formal education than other Democratic-oriented groups. They are the group with the largest share of Black non-Hispanic adults (26\%), and six-in-ten are women. They are the only Democratic-oriented typology group in which a larger share say that the decline in the share of Americans belonging to an organized religion is bad for society than say this is good for society.

\xhdr{Establishment Liberal.}
Holding liberal positions on nearly all issues, Establishment Liberals are some of the strongest supporters of the current president and the Democratic Party of any political typology group. While deeply liberal, Establishment Liberals are the typology group most likely to see value in political compromise and tend to be more inclined toward more measured approaches to societal change than their Progressive Left counterparts. Like other Democratic-oriented groups, most (73\%) say a lot more needs to be done to ensure racial equality. Yet they are the only Democratic-aligned group in which a majority of those who say a lot more needs to be done also say this can be achieved by working within the current system. About half (51\%) are White, while 18\% are Black, 20\% are Hispanic and 10\% are Asian. Nearly nine-in-ten (89\%) say that compromise is how things get done in politics, higher than the share in any other political typology group.

\xhdr{Progressive Left.}
Reflecting their name, Progressive Left have very liberal views across a range of issues---including the size and scope of government, foreign policy, immigration and race. A sizable majority (79\%) describe their views as liberal, including 42\% who say their views are very liberal. Roughly two-thirds (68\%) are White, non-Hispanic, by far the largest share among Democratic-aligned groups. Progressive Left are the second youngest typology group---71\% are ages 18 to 49. They are also highly educated, with about half (48\%) holding at least a four-year college degree. Their views on race and racial equality distinguish them from other typology groups: sizable majorities say White people benefit from societal advantages that Black people do not have and that most U.S.\ institutions need to be completely rebuilt to ensure equal rights for all Americans regardless of race or ethnicity.

\vspace{1mm}
\noindent 
When we stratify personas by political leaning, we group \emph{right-leaning} personas as those assigned Faith and Flag Conservative, Committed Conservative, Populist Right, or Ambivalent Right, and \emph{left-leaning} personas as those assigned Outsider Left, Democratic Mainstay, Establishment Liberal, or Progressive Left. 
Stressed Sideliners constitute a separate, politically disengaged category; however, no persona in our final sample is assigned this typology (see Table~\ref{tab:persona-invariant}).

\subsection{Behavior Policy}
\label{sec:policy}
Each sock puppet's behavior is governed by a \emph{persona prompt} that conditions the LLM on a
specific demographic and political profile:
  
\begin{promptbox}
\textbf{Persona Prompt} \\
Imagine that you \texttt{\{gender\}}. You are \texttt{\{age\}} years old and 
live \texttt{\{preposition\_location\}} \texttt{\{location\}}. You are 
\texttt{\{race\}}, and \texttt{\{education\}}. You identify politically as 
\texttt{\{political\_type\}}, which is described by Pew as follows: 
\texttt{\{poltype\_description\}}.
\end{promptbox}

Each placeholder in the persona prompt is populated as follows:

\begin{itemize}
    \item \texttt{\{gender\}}: One of ``identify as a man,'' ``identify as a woman,'' or ``do not identify as a man or a woman.''

    \item \texttt{\{age\}}: A randomly generated integer within the sampled age bracket (18--29, 30--49, 50--64, or 65--85).

    \item \texttt{\{preposition\_location\}}: Either ``in'' or ``near,'' selected uniformly at random.

    \item \texttt{\{location\}}: A U.S.\ city sampled from incorporated places with population of 50,000 or more, weighted by population.

    \item \texttt{\{race\}}: One of ``White and Non-Hispanic,'' ``White and Hispanic,'' ``Black,'' ``Asian,'' ``American Indian,'' ``Native Hawaiian and Other Pacific Islander,'' or ``Multiracial.''

    \item \texttt{\{education\}}: Highest attainment level (see Table~\ref{tab:education}).

    \item \texttt{\{political\_type\}}: One of the nine categories from Pew's Political Typology: ``Faith
  and Flag Conservative,'' ``Committed Conservative,'' ``Populist Right,'' ``Ambivalent Right,''
  ``Stressed Sideliner,'' ``Outsider Left,'' ``Democratic Mainstay,'' ``Establishment Liberal,'' or
  ``Progressive Left.''

    \item \texttt{\{poltype\_description\}}: A detailed description of the political typology category drawn from Pew Research Center's published characterizations~\citep{pew2021typology}. Full descriptions are provided in Section~\ref{sec:pol}.
    
\end{itemize}

\begin{table}[ht]
  \begin{tabular}{l}
  \toprule
  Prompt phrase \\
  \midrule
  ``you received no education'' \\
  ``your highest level of education is [4th/6th/8th/9th/10th/11th] grade'' \\
  ``you graduated high school and did not go to college'' \\
  ``you received some college education and did not finish your degree'' \\
  ``your highest level of education is an [occupational/academic] Associate's degree'' \\
  ``your highest level of education is a [Bachelor's/Master's/professional/Doctoral] degree'' \\
  \bottomrule
  \end{tabular}
  \vspace{0.2em}
  \caption{Education level prompts.}
  \label{tab:education}
  \end{table}

\subsection{Platform-Visible Signals}
\label{sec:signals}
Each counterfactual condition perturbs exactly one demographic dimension while holding the behavioral policy fixed. We target location, age, and gender because $\Xtwitter$ plausibly observes or infers each: location via IP geolocation, age via the declared birth date at registration, and gender via the display name. On other platforms, different signals may be observable and thus amenable to perturbation. Below we describe how each perturbed signal reaches the platform in our deployment.

\xhdr{Location ($C=1$).}
The platform does not ask users to declare their location; instead, it infers geographic information from the account's IP address. We route each account's traffic through a residential proxy service (SmartProxy) that provides state-level sticky sessions. During preprocessing, we query the provider's API
for state-specific port ranges and assign each account a unique port within the range corresponding to its assigned state. For baseline and non-location conditions ($C \in \{0,2,3\}$), the assigned state is the persona's baseline location; for location-perturbed accounts ($C=1$), it is the counterfactual state. At runtime, each account's browser connects through \texttt{state.smartproxy.com} on its assigned port, so the platform observes a residential IP in the target state for every session. 

\xhdr{Age ($C=2$).}
The platform requires a birth date during account registration. For each account, we generate a random birth date within the account's assigned age bracket: for baseline and non-age conditions ($C \in \{0,1,3\}$) this is the persona's baseline bracket, while for age-perturbed accounts ($C=2$) it is the
counterfactual bracket. The birth date is entered into the platform's registration form, providing the primary channel through which the platform observes the account's age.

\xhdr{Gender ($C=3$).}
\label{sec:name_gen}
Gender is communicated to the platform through the account's display name. We generate names conditional on demographic attributes using GPT-4o via the following prompt:

  \begin{promptbox}
  \textbf{Name Generation Prompt} \\
  \texttt{\{demographic\_prompt\}} Please consider 200 unique full names that would likely fit this person
  then randomly choose 100. Please return the 100 that you choose as an array of strings. Please do not
  write anything else in your response, just the array of strings of names. The output should follow the
   form ``[name1, name2, ...]'' and contain nothing else.
  \end{promptbox}

The prompt \texttt{\{demographic\_prompt\}} follows the template defined in Section~\ref{sec:policy} but with the gender field set to match the account's \emph{assigned condition} rather than the baseline. We generate a
pool of 100 candidate names per persona---larger than the 80 accounts per persona---and then randomly
sample without replacement when assigning names to individual accounts.

\section{Details on Deployment and Interaction}
\subsection{Account Distribution}
We initially planned to deploy 1,120 sock puppets -- 80 replicas for 14 different personas. However, due to verification failures at account creation time, 826 accounts were ultimately deployed successfully, yielding 9,733 user-day observations. 

\begin{figure}[htbp]
      \centering
      \begin{subfigure}[b]{\textwidth}
          \centering
          \includegraphics[width=0.7\textwidth]{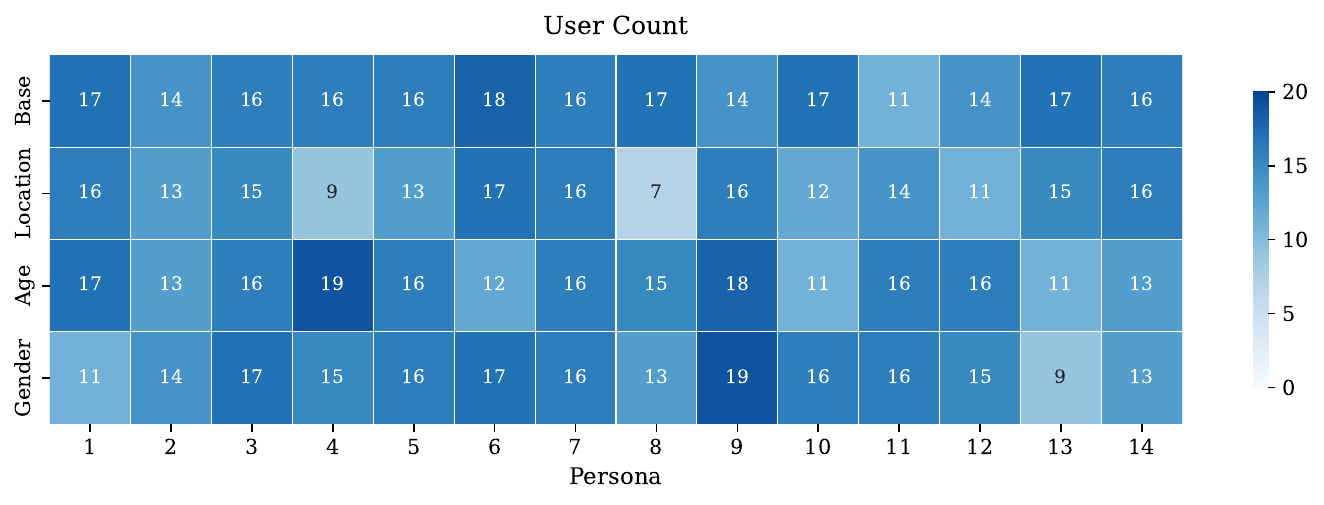}
          \caption{Number of active users per cell.}
          \label{fig:cell_sizes}
      \end{subfigure}

      \vspace{0.5cm}

      \begin{subfigure}[b]{\textwidth}
          \centering
          \includegraphics[width=0.7\textwidth]{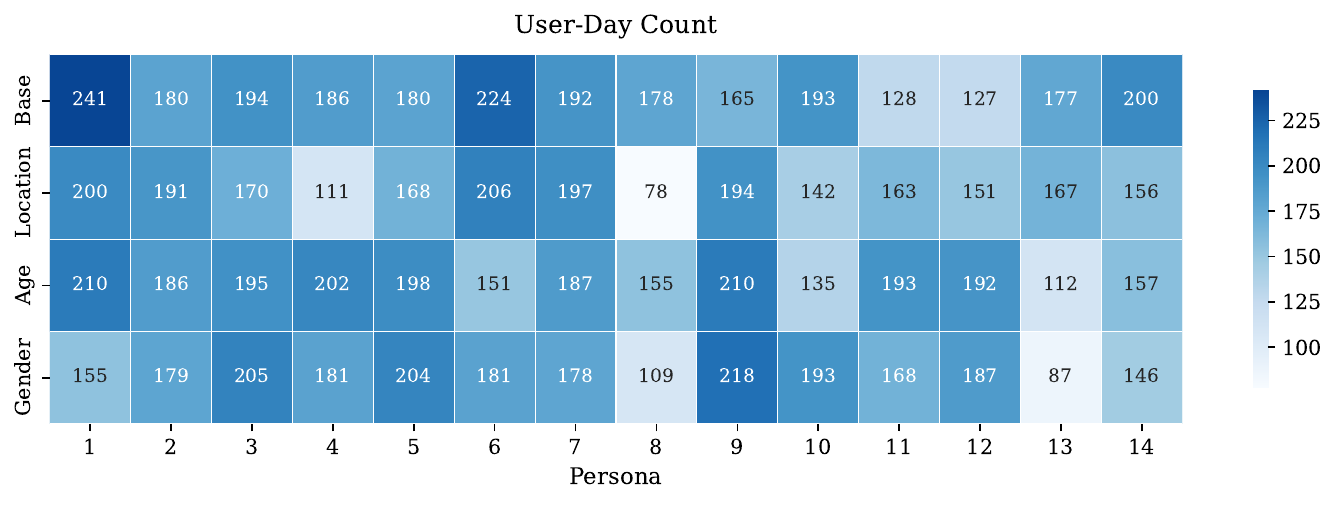}
          \caption{Number of user-day observations per cell.}
          \label{fig:observations}
      \end{subfigure}
      \caption{Sample sizes by persona $\times$ treatment condition.}
      \label{fig:sample_sizes}
  \end{figure}

Table~\ref{tab:sample_summary} summarizes the distribution of observation time across accounts. The
  majority of accounts contributed between 10 and 16 days of data, with 9.7\% achieving near-full
  coverage and 2.3\% experiencing early dropout. We also report the distribution of user and user--day observations stratified by counterfactual category in Figures \ref{fig:cell_sizes} and \ref{fig:observations}.

  \begin{table}[htbp]
  \begin{tabular}{lc}
  \toprule
  Statistic & Value \\
  \midrule
  Accounts successfully deployed & 826 \\
  User-day observations & 9,733 \\
  \addlinespace
  \multicolumn{2}{l}{\textit{Days active per account}} \\
  \quad Mean & 13.2 days \\
  \quad Median & 12 days \\
  \quad IQR & [10, 16] days \\
  \quad Maximum & 23 days \\
  \addlinespace
  Near-full coverage ($\geq 21$ days) & 9.7\% \\
  Early dropout ($\leq 3$ days) & 2.3\% \\
  \bottomrule
  \end{tabular}
  \vspace{0.2em}
    \caption{Sample Summary}
  \label{tab:sample_summary}
  \end{table}

\subsection{Feed Engagement}
\label{sec:feed_engagement}

Each sock puppet interacts with the platform's feed through an LLM-mediated engagement loop. The agent observes tweets served by the recommendation algorithm and decides how to respond based on its persona. 
We define four possible actions an agent can take in response to each tweet:

  \begin{itemize}
      \item \textbf{Like}: Signal positive engagement with the content.
      \item \textbf{Follow}: Subscribe to the tweet's author for future content.
      \item \textbf{Read Replies}: View the conversation thread, signaling interest without explicit endorsement (selecting this action navigates into the reply thread, where up to five replies are recursively processed through the same pipeline before returning to the main feed).
      \item \textbf{Ignore}: Skip the tweet without interaction.
  \end{itemize}

These actions mirror the primary engagement mechanisms available to human users on the platform. To ensure sufficient engagement signal for the recommendation algorithm to learn user preferences, we require agents to take at least one active action (like, follow, or read replies) every five tweets. This constraint prevents purely passive scrolling behavior that would provide minimal signal to the algorithm. 
Importantly, the constraint is specified only as a prompt-level instruction: each content item is processed in a single, stateless LLM call with no access to prior actions. The instruction therefore acts as a behavioral nudge rather than a hard constraint. 

The engagement behavior is controlled by two prompts. The \emph{system prompt} establishes the agent's
identity and explains the available actions:

\begin{promptbox}
\textbf{Feed Engagement Prompt (System Prompt)} \\
Welcome to Twitter! In future messages, you will see a sequence of tweets and have to decide how to react to each one. 

\noindent \{persona\_prompt\}

\noindent In our conversation below, I will show you a sequence of Tweets on Twitter. Please use your persona as described above to decide how you would like to react to each one.

\noindent You are primarily on Twitter to read tweets, but you can also follow accounts, like tweets, and read replies to tweets.

\noindent Below, I list the possible actions you can take.
\begin{itemize}
   \item If you want to see more tweets like the one you just saw, you can follow the author (to do this, include "[FOLLOW]" in your response).
   \item If you are curious to see what other people think about the tweet, you can read the replies (to do this, include "[READ-REPLY]" in your response).
   \item If you like the tweet, you can like it (to do this, include "[LIKE]" in your response), 
   \item If you do not want to react to a tweet (for example, if you are not interested in it), you can ignore it (to do this, include "[IGNORE]" in your response).
\end{itemize}

\noindent You should use one of the first three actions ([FOLLOW], [READ-REPLY], [LIKE]) at least once every 5 tweets. You can use [IGNORE] as many times as you like.

\noindent Your response should contain either [IGNORE], [FOLLOW], [READ-REPLY], or [LIKE]. If it contains [IGNORE], please do not include any other reactions.
\end{promptbox}

For each tweet encountered, the agent receives a \emph{user prompt} containing the tweet's
content and metadata (which optionally contains the quoted tweet text and media image if present):

\begin{promptbox}
\textbf{Tweet Prompt (User Prompt)} \\
You are on twitter and see the following tweet - please first repeat the tweet to ensure that you have understood it, then explain your reaction.

\noindent Author: \{author\}

\noindent Tweet Text: \{tweetText\}

\noindent Likes: \{likes\}

\noindent Retweets: \{retweets\}

\noindent Replies: \{replies\}

\end{promptbox}

We include engagement metrics (likes, retweets, replies) to provide social proof signals
that human users would observe. The instruction to ``repeat the tweet'' serves as a chain-of-thought
mechanism, ensuring the agent processes the content before deciding on an action. The agent's chosen
action is then executed on the platform via automated browser interactions.

\subsection{Action Distribution}
  \label{sec:action_distribution}

Table~\ref{tab:actions} reports the distribution of actions across all tweet interactions. Agents ignore or read replies to the vast majority of tweets. Likes are relatively rare, and follows are near-absent.

      \begin{table}[htbp]
    \centering
    \caption{Distribution of agent actions.}
    \label{tab:actions}
    \begin{tabular}{lrrrrr}
    \toprule
     & Read-Reply & Ignore & Like & Follow & Total \\
    \midrule
    Count     & 139,000 & 90,151 & 14,095 & 13 & 243,259 \\
    Share (\%) & 57.1 & 37.1 & 5.8 & $<0.01$ & \\
    \bottomrule
    \end{tabular}
    \end{table}

  In practice, agents engage far more often than the 20\% floor implied by the one-in-five nudge: they take an active action (Read-Reply, Like, or Follow) on approximately 63\% of rendered tweets. We also examined the gaps between consecutive active actions within each account--feed stream: only 1.17\% of these gaps contain five or more consecutive Ignores, confirming that the soft rule is respected in the vast majority of cases.

Engagement rates are balanced across treatment arms: Base 63.7\%, Location-CF 62.7\%, Age-CF 62.5\%, and Gender-CF 62.9\%. A chi-square test on the full action-by-condition table yields Cram\'er's $V = 0.009$, well below Cohen's 0.07 threshold for a small effect, so any cross-condition difference in engagement is negligible.

 \subsection{Attrition Analysis}
  \label{sec:attrition}

  Our identification strategy for the counterfactual analysis relies on random assignment of treatment conditions within each persona.
  If accounts in certain treatment conditions are more likely to be banned or suspended during the experiment, the resulting imbalance could bias our estimates. Therefore, we test whether, conditional on
  successful deployment, observation time varies systematically by treatment. 

  Note that we focus on mid-experiment attrition (bans and suspensions) rather than pre-deployment failures at the account creation stage. Unlike in human experiments, pre-deployment attrition cannot introduce
  selection bias in our setting: accounts within the same persona--treatment cell are exchangeable replicas running an identical behavioral policy, so creation failures only reduce statistical power, not the
  representativeness of surviving accounts.

  Table~\ref{tab:days_by_treatment} reports observation time by treatment condition among the 826
  deployed accounts. To test for differential attrition, we regress days active on treatment indicators with fixed
  effects:
  \begin{equation}
  \text{DaysActive}_i = \alpha + \beta_{\text{loc}} \cdot \mathbf{1}[\text{Location}] +
  \beta_{\text{age}} \cdot \mathbf{1}[\text{Age}] + \beta_{\text{gen}} \cdot \mathbf{1}[\text{Gender}] +
   \gamma_{p_i} + \varepsilon_i
  \end{equation}

No treatment coefficient is statistically significant (Location: $p = 0.805$; Age: $p = 0.924$;
Gender: $p = 0.301$), indicating that observation time is balanced across conditions. Therefore, we find no evidence that treatment assignment affects the likelihood of mid-experiment attrition.

  \begin{table}[htbp]
  \centering
  \caption{Observation Time by Treatment Condition}
  \label{tab:days_by_treatment}
  \begin{tabular}{lcccc}
  \toprule
  Treatment & $N$ & Mean & Median & IQR \\
  \midrule
  Base & 219 & 13.20 & 12 & [10, 16] \\
  Location & 191 & 13.39 & 12 & [10, 17] \\
  Age & 209 & 13.30 & 13 & [10, 16] \\
  Gender & 207 & 12.80 & 12 & [10, 16] \\
  \bottomrule
  \end{tabular}
  \end{table}

\section{Tweet Labeling}
  \label{sec:labels}

  We classify tweet content along two dimensions: political content and polarizing content. Both
  classifiers use gpt-oss-20b \cite{openai2025gptoss120bgptoss20bmodel}, with
  chain-of-thought prompting and few-shot learning ($k=4$ examples selected via bootstrap optimization).

  \subsection{Political Content}
  \label{sec:political-labeling}

  We classify each tweet into one of four mutually exclusive categories: \texttt{not\_political},
  \texttt{political\_left}, \texttt{political\_right}, or \texttt{political\_neutral}. The classifier is
   prompted as follows:

\begin{promptbox}
\textbf{Political Classification Prompt}

  \noindent The tweet was posted in November 2024, during the U.S. presidential election period.

  \noindent Classify (1) whether the tweet contains substantive U.S. political content, and (2) if
  political, its ideological alignment in U.S. politics.

  \noindent Definition 1 --- Political vs Not Political

  \noindent Label a tweet as ``political'' only if it clearly refers to at least one of the following:

  \noindent - The U.S. presidential election (campaigning, voting, results, legitimacy, turnout)

  \noindent - U.S. political institutions, public policy, or governance

  \noindent - Formal political actors (candidates, elected officials, political parties)

  \noindent - Explicit partisan or electoral positioning

  \noindent If it is unclear whether the tweet contains substantive political content, label it
  ``not\_political''.

  \noindent Definition 2 --- Ideological Alignment (only if political)

  \noindent Classify ideological alignment based only on the political or policy position explicitly
  expressed or endorsed by the tweet.

  \noindent - ``political\_left'': Explicit endorsement of U.S. liberal or progressive positions or
  policies

  \noindent - ``political\_right'': Explicit endorsement of U.S. conservative positions or policies

  \noindent - ``political\_neutral'': Political content that is informational, mixed, anti-partisan, or
  lacks a clear ideological endorsement

  \noindent Important Guidelines:

  \noindent - Ideological alignment may be inferred from praise, criticism, or sarcasm only when the
  tweet expresses a clear and unambiguous political stance

  \noindent - Emotional reactions or election outcome commentary without a clear stance should be
  labeled ``political\_neutral''

  \noindent - Mentions of political figures alone are not sufficient for left/right classification

  \noindent Output exactly one of: ``not\_political'', ``political\_left'', ``political\_right'',
  ``political\_neutral''.
  \end{promptbox}

For analysis, we derive three outcome variables from these labels:
  \begin{itemize}
      \item Politicalness: Binary indicator for whether the tweet is political (i.e., not
  \texttt{not\_political}).
      \item Right-leaning: Binary indicator for \texttt{political\_right}, computed over all
  tweets.
      \item Left-leaning: Binary indicator for \texttt{political\_left}, computed over all
  tweets.
  \end{itemize}

  \subsection{Polarizing Content}
  \label{sec:polarization-labeling}

  We classify each tweet into one of six mutually exclusive categories, according to \cite{naseem2025polarbenchmarkmultilingualmulticultural}: \texttt{not\_polarizing},
  \texttt{polarizing\_political}, \texttt{polarizing\_religious}, \texttt{polarizing\_sexual},
  \texttt{polarizing\_racial}, or \\\texttt{polarizing\_other}. The classifier is prompted as follows:

  \begin{promptbox}
  \textbf{Polarization Classification Prompt}

  \noindent You will be given a tweet collected in November 2024, during the U.S. presidential election
  period. Your task is to determine whether it is polarizing and, only if so, identify the primary type
  of polarization.

  \noindent Definition:

  \noindent A tweet is polarizing if and only if it satisfies both conditions:

  \noindent 1. It distinguishes between two or more social, political, religious, sexual, or racial
  groups (explicitly or implicitly), and

  \noindent 2. It frames those groups in a moralized conflict, such as: blame or hostility; moral
  superiority or inferiority; threat or danger; victim-oppressor narratives.

  \noindent If either condition is not met, the tweet is not polarizing.

  \noindent Important Constraints

  \noindent - Mentioning politics, religion, gender, sexuality, race, or ideology alone is not
  sufficient for polarization.

  \noindent - Statements, slogans, values, prayers, or opinions without group-based blame or moral
  conflict must be labeled ``not\_polarizing''.

  \noindent - Emotional language is irrelevant unless it is directed at a group as a group.

  \noindent - Attacks on individuals are polarizing only if they generalize to a broader group.

  \noindent - Intra-group discussion or disagreement is not polarizing unless subgroups are framed as
  enemies, traitors, or morally inferior.

  \noindent Output exactly one of: ``not\_polarizing'', ``polarizing\_political'',
  ``polarizing\_religious'', ``polarizing\_sexual'', ``polarizing\_racial'', ``polarizing\_other''.
  \end{promptbox}

For analysis, due to the sparsity of the polarizing categories, we collapse the six categories into a binary {polarization} indicator equal to 1
if the tweet is classified into any polarizing category, and 0 otherwise.

\section{Robustness of Amplification}
  \label{sec:amplification_more}

This appendix presents robustness checks for the amplification ratio analysis in Section~\ref{sec:amplification}. We examine whether estimates are sensitive to the exposure window used for aggregation (that is, whether using only early observations versus the full deployment period produces different amplification ratios). A key identifying assumption of our estimator is that browsing one feed does not causally affect content shown in the other. Potential violations (such as engagement spillovers or preference updates across feeds) would make feeds more similar over time, biasing amplification estimates toward zero. If such cross-feed contamination occurs, we would expect amplification estimates to \emph{decline} as accounts accumulate browsing history, since the algorithm would increasingly tailor both feeds to the same learned preferences.

To address this, we compute amplification ratios using cumulative exposure windows of increasing
length: each account's first day of activity only, first 7 days (Week~1), first 14 days (Week~2), and the full deployment period. Table~\ref{tab:robustness-cumulative} reports amplification estimates for each content type across
these windows.

\begin{table}[htbp]
    \centering
    \small
    \begin{tabular}{l cccc}
    \toprule
    \textbf{Content} & \textbf{Day 1} & \textbf{Week 1} & \textbf{Week 2} & \textbf{Full} \\
    \midrule
    Political
      & $+12.7\%$ {\scriptsize$[+7.8,\, +18.1]$}
      & $+14.6\%$ {\scriptsize$[+9.5,\, +20.2]$}
      & $+14.2\%$ {\scriptsize$[+9.1,\, +19.9]$}
      & $+14.2\%$ {\scriptsize$[+9.1,\, +19.8]$} \\[2pt]
    Right-Leaning
      & $+29.8\%$ {\scriptsize$[+17.4,\, +43.0]$}
      & $+24.2\%$ {\scriptsize$[+16.2,\, +33.4]$}
      & $+23.4\%$ {\scriptsize$[+15.8,\, +32.1]$}
      & $+23.3\%$ {\scriptsize$[+15.8,\, +31.9]$} \\[2pt]
    Left-Leaning
      & $-21.7\%$ {\scriptsize$[-30.0,\, -13.2]$}
      & $-6.2\%$  {\scriptsize$[-14.1,\, +1.4]$}
      & $-6.1\%$  {\scriptsize$[-14.6,\, +2.2]$}
      & $-6.1\%$  {\scriptsize$[-14.6,\, +2.1]$} \\[2pt]
    Toxic
      & $+37.9\%$ {\scriptsize$[+9.9,\, +75.3]$}
      & $+42.5\%$ {\scriptsize$[+16.9,\, +74.5]$}
      & $+39.4\%$ {\scriptsize$[+14.5,\, +69.9]$}
      & $+39.2\%$ {\scriptsize$[+14.4,\, +69.7]$} \\[2pt]
    Polarizing
      & $+48.4\%$ {\scriptsize$[+35.2,\, +63.1]$}
      & $+32.8\%$ {\scriptsize$[+24.2,\, +42.3]$}
      & $+32.6\%$ {\scriptsize$[+24.1,\, +42.0]$}
      & $+32.3\%$ {\scriptsize$[+23.7,\, +41.6]$} \\[2pt]
    High-reach
      & $+14.5\%$ {\scriptsize$[+0.8,\, +35.5]$}
      & $+3.3\%$  {\scriptsize$[-6.3,\, +16.8]$}
      & $+3.4\%$  {\scriptsize$[-6.1,\, +16.6]$}
      & $+3.4\%$  {\scriptsize$[-6.1,\, +16.7]$} \\
    \bottomrule
    \end{tabular}
    \vspace{0.3em}
    \caption{Amplification estimates by cumulative exposure window. Uncertainty
     is 95\% CI from persona-level cluster bootstrap: in each iteration, we resample personas with
  replacement and recompute the amplification ratio, preserving the within-persona correlation structure.}
    \label{tab:robustness-cumulative}
    \end{table}

We observe that political and toxic content show no statistically significant change across exposure windows, with confidence intervals overlapping substantially throughout. Right-leaning, polarizing, and high-reach content exhibit attenuation from Day~1 to Week~1 (29.8\% to 24.2\%, 48.4\% to 32.8\%, and 14.5\% to 3.3\%, respectively), followed by stabilization. 
The attenuation is sharpest for high-reach content, which is amplified on Day~1 ($+14.5\%$, significant) but drops to a small and non-significant effect by Week~1 ($+3.3\%$), where it stabilizes. Similarly, left-leaning content shows Day~1 de-amplification ($-$21.7\%, significant) that attenuates toward zero by Week~1 ($-$6.2\%, no longer significant) and stabilizes thereafter.
  
The early attenuation observed for some content types (Day~1 to Week~1) likely reflects the algorithm
calibrating to user behavior before reaching steady state. Crucially, estimates stabilize by Week~1
and remain unchanged thereafter. If cross-feed spillovers were substantially biasing estimates toward
zero, we would expect continued attenuation throughout the deployment; instead, the estimates plateau.
This stability supports our identifying assumption that browsing one feed does not meaningfully affect content shown in the other. Moreover, to the extent that any residual bias exists, it would attenuate estimates toward zero, implying that our reported amplification effects are, if anything, conservative.

\section{Pooled Treatment Effects}
\label{sec:pooled-robustness}
We also report pooled treatment effects estimated without control variables as a robustness check. Results are nearly identical to the primary specification (Table~\ref{tab:pooled-ate}), confirming that
  the controls do not affect the estimates.

\begin{table}[htbp]
  \centering
  \small
  \begin{tabular}{l cc cc cc}
  \toprule
  & \multicolumn{2}{c}{\textbf{Location} ($D^{\text{urb}}$)}
  & \multicolumn{2}{c}{\textbf{Age} ($D^{\text{age}}$)}
  & \multicolumn{2}{c}{\textbf{Gender} ($D^{\text{male}}$)} \\
  \cmidrule(lr){2-3} \cmidrule(lr){4-5} \cmidrule(lr){6-7}
  \textbf{Content}
  & $\hat{\beta}$ & \textbf{95\% CI}
  & $\hat{\beta}$ & \textbf{95\% CI}
  & $\hat{\beta}$ & \textbf{95\% CI} \\
  \midrule
  Political
    & $-1.70$ & {\scriptsize$[-3.86,\,0.46]$}
    & $0.70$  & {\scriptsize$[-1.26,\,2.66]$}
    & $1.00$  & {\scriptsize$[-1.16,\,3.16]$} \\
    & {\scriptsize$(0.137)$} &
    & {\scriptsize$(0.501)$} &
    & {\scriptsize$(0.408)$} & \\[2pt]

  Right-leaning
    & $-0.50$ & {\scriptsize$[-1.87,\,0.87]$}
    & $0.90$  & {\scriptsize$[-0.28,\,2.08]$}
    & $-0.40$ & {\scriptsize$[-1.58,\,0.78]$} \\
    & {\scriptsize$(0.440)$} &
    & {\scriptsize$(0.111)$} &
    & {\scriptsize$(0.561)$} & \\[2pt]

  Left-leaning
    & $-0.60$ & {\scriptsize$[-1.38,\,0.18]$}
    & $-0.30$ & {\scriptsize$[-1.08,\,0.48]$}
    & $0.40$  & {\scriptsize$[-0.38,\,1.18]$} \\
    & {\scriptsize$(0.183)$} &
    & {\scriptsize$(0.383)$} &
    & {\scriptsize$(0.379)$} & \\[2pt]

  Toxic
    & $0.10$           & {\scriptsize$[-0.68,\,0.88]$}
    & $\mathbf{0.80}$  & {\scriptsize$\mathbf{[0.02,\,1.58]}$}
    & $-0.10$          & {\scriptsize$[-0.69,\,0.49]$} \\
    & {\scriptsize$(0.831)$} &
    & {\scriptsize$\mathbf{(0.033)}$} &
    & {\scriptsize$(0.868)$} & \\[2pt]

  Polarizing
    & $\mathbf{-1.60}$ & {\scriptsize$\mathbf{[-2.97,\,-0.23]}$}
    & $0.50$           & {\scriptsize$[-0.68,\,1.68]$}
    & $-0.70$          & {\scriptsize$[-2.07,\,0.67]$} \\
    & {\scriptsize$\mathbf{(0.020)}$} &
    & {\scriptsize$(0.444)$} &
    & {\scriptsize$(0.304)$} & \\[2pt]

  High-reach
    & $1.10$  & {\scriptsize$[-1.45,\,3.65]$}
    & $-0.90$ & {\scriptsize$[-3.64,\,1.84]$}
    & $-0.10$ & {\scriptsize$[-2.84,\,2.64]$} \\
    & {\scriptsize$(0.381)$} &
    & {\scriptsize$(0.510)$} &
    & {\scriptsize$(0.928)$} & \\
  \bottomrule
  \end{tabular}

  \begin{minipage}{\columnwidth}
  \vspace{0.5em}
    \caption{Pooled treatment effects on algorithmic lift, without control variables. Each
   coefficient represents the average effect of a one-unit change in the demographic signal on the gap
  between \emph{For You} and \emph{Following} content rates. Coefficients for binary outcomes are scaled
  by 100 and expressed in percentage points. Values in parentheses are $p$-values
  computed from standard errors clustered at the account level. Significant coefficients ($p < 0.05$) shown in bold.}
  \label{tab:pooled-ate-nocontrols}
  \end{minipage}
\end{table}

\section{Heterogeneous Treatment Effects}
\label{sec:heterogeneous-ate}

This appendix reports persona-specific treatment effects on algorithmic lift for all six outcomes, estimated via model~\eqref{eq:het_reg}. Each figure shows the estimated coefficient and 95\% confidence interval for each persona--treatment pair. Filled markers indicate significance at the 5\% level.

    \begin{figure}[htbp]
        \centering
        \includegraphics[width=\textwidth]{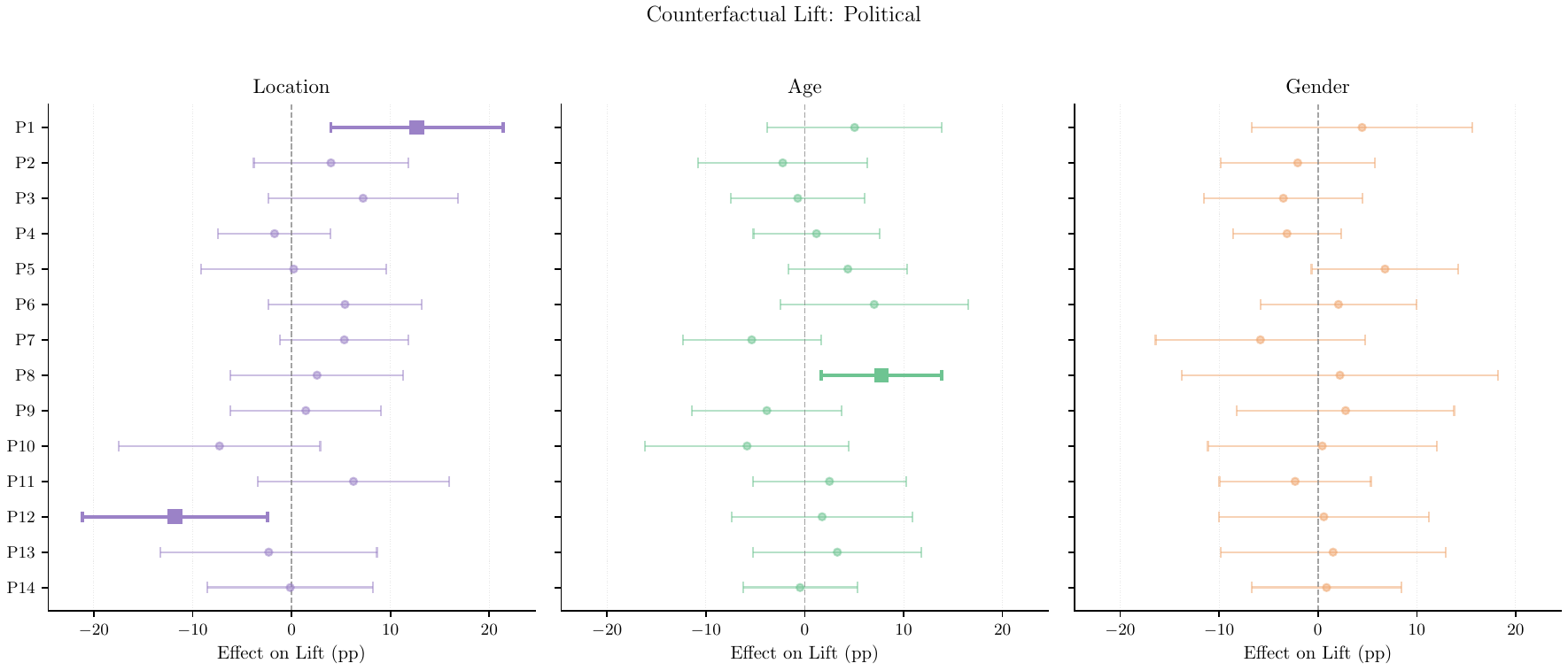}
        \caption{Persona-specific treatment effects on algorithmic lift for political content. Each row is a persona; each panel shows a different treatment (location, age, gender). 95\% confidence intervals.}
        \label{fig:ate_politicalness}
    \end{figure}

    \begin{figure}[htbp]
        \centering
        \includegraphics[width=\textwidth]{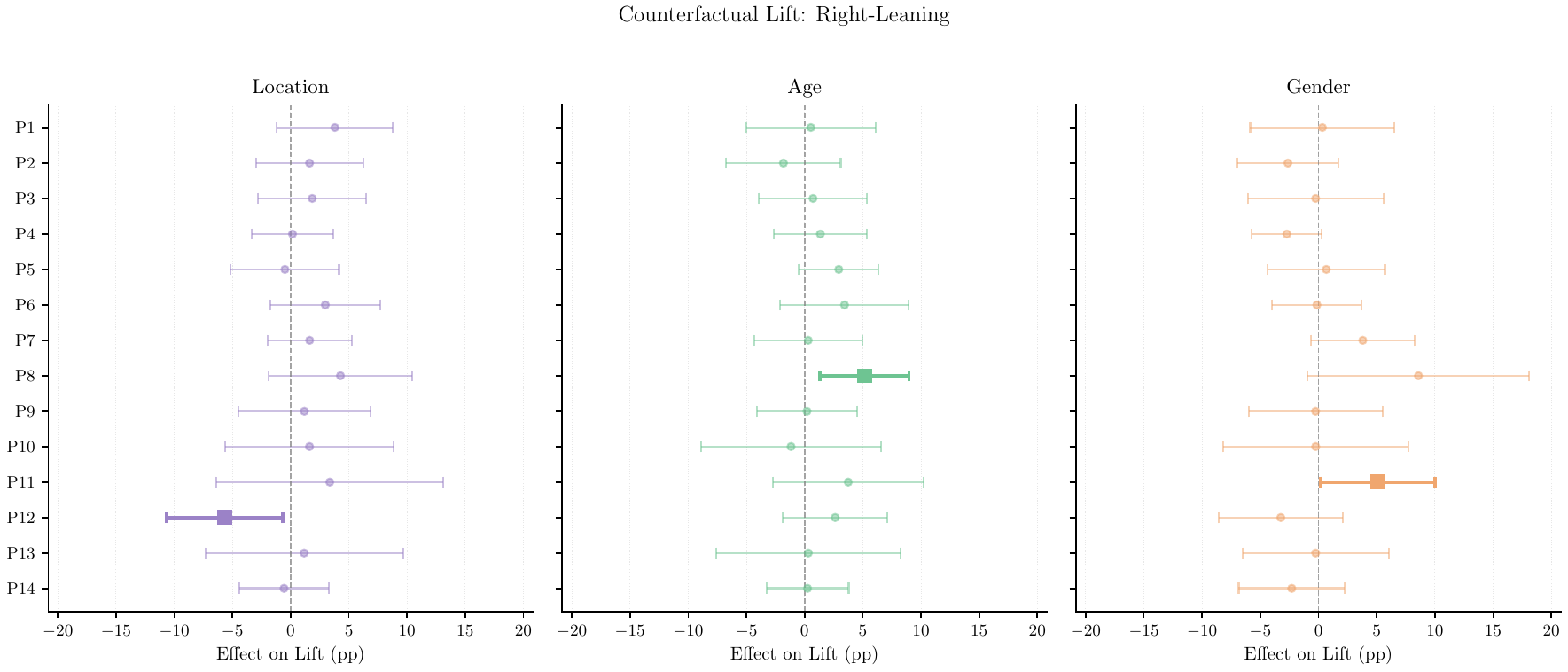}
        \caption{Persona-specific treatment effects on algorithmic lift for right-leaning content. Each row is a persona; each panel shows a different treatment (location, age, gender). 95\% confidence intervals.}
        \label{fig:ate_right_leaning}
    \end{figure}

    \begin{figure}[htbp]
        \centering
        \includegraphics[width=\textwidth]{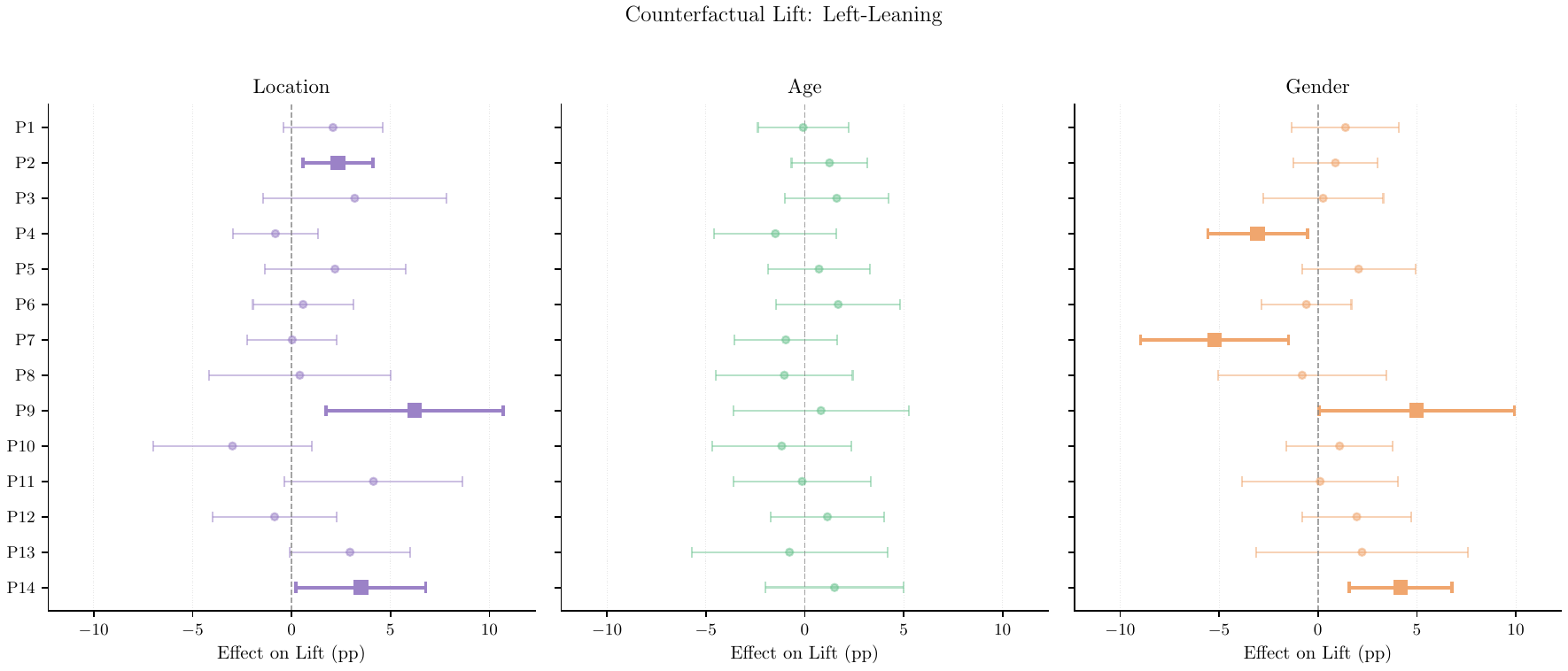}
        \caption{Persona-specific treatment effects on algorithmic lift for left-leaning content. Each row is a persona; each panel shows a different treatment (location, age, gender). 95\% confidence intervals.}
        \label{fig:ate_left_leaning}
    \end{figure}

    \begin{figure}[htbp]
        \centering
        \includegraphics[width=\textwidth]{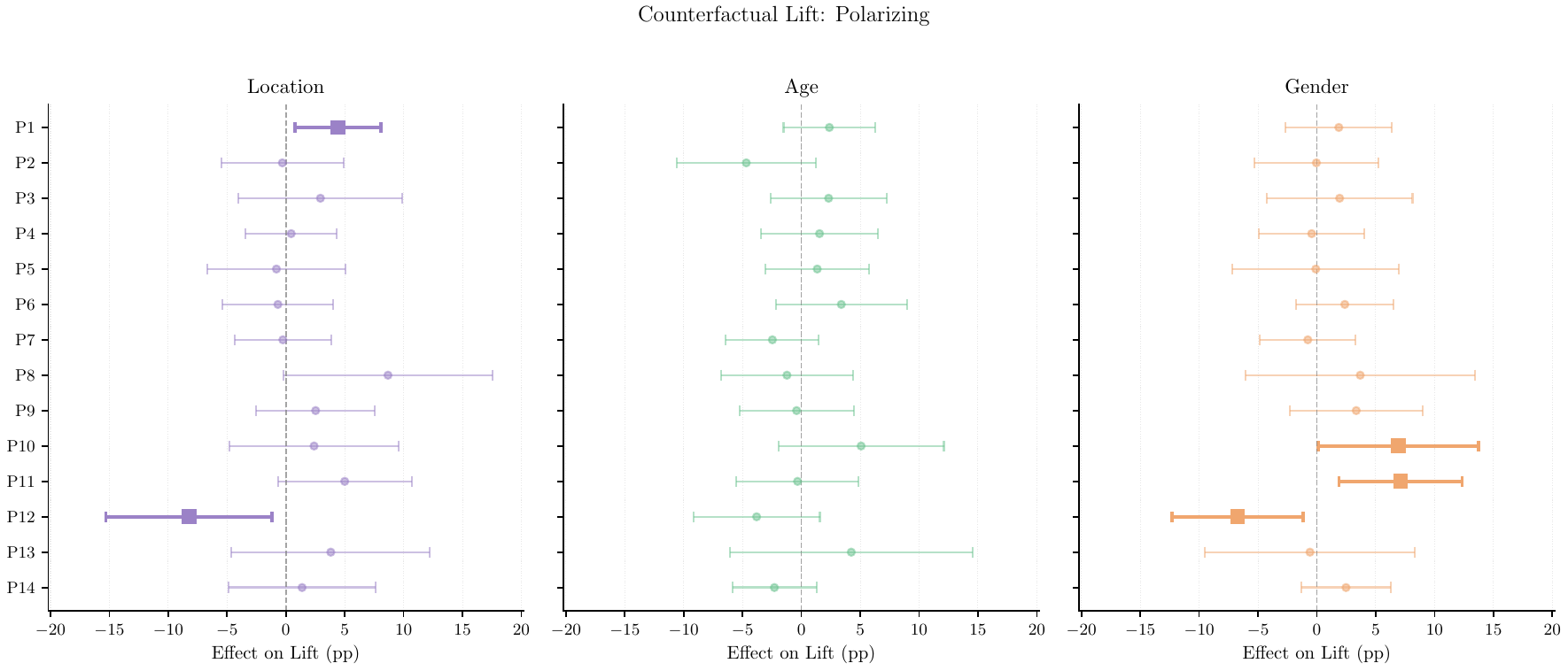}
        \caption{Persona-specific treatment effects on algorithmic lift for polarizing content. Each row is a persona; each panel shows a different treatment (location, age, gender). 95\% confidence intervals.}
        \label{fig:ate_polarization}
    \end{figure}

    \begin{figure}[htbp]
        \centering
        \includegraphics[width=\textwidth]{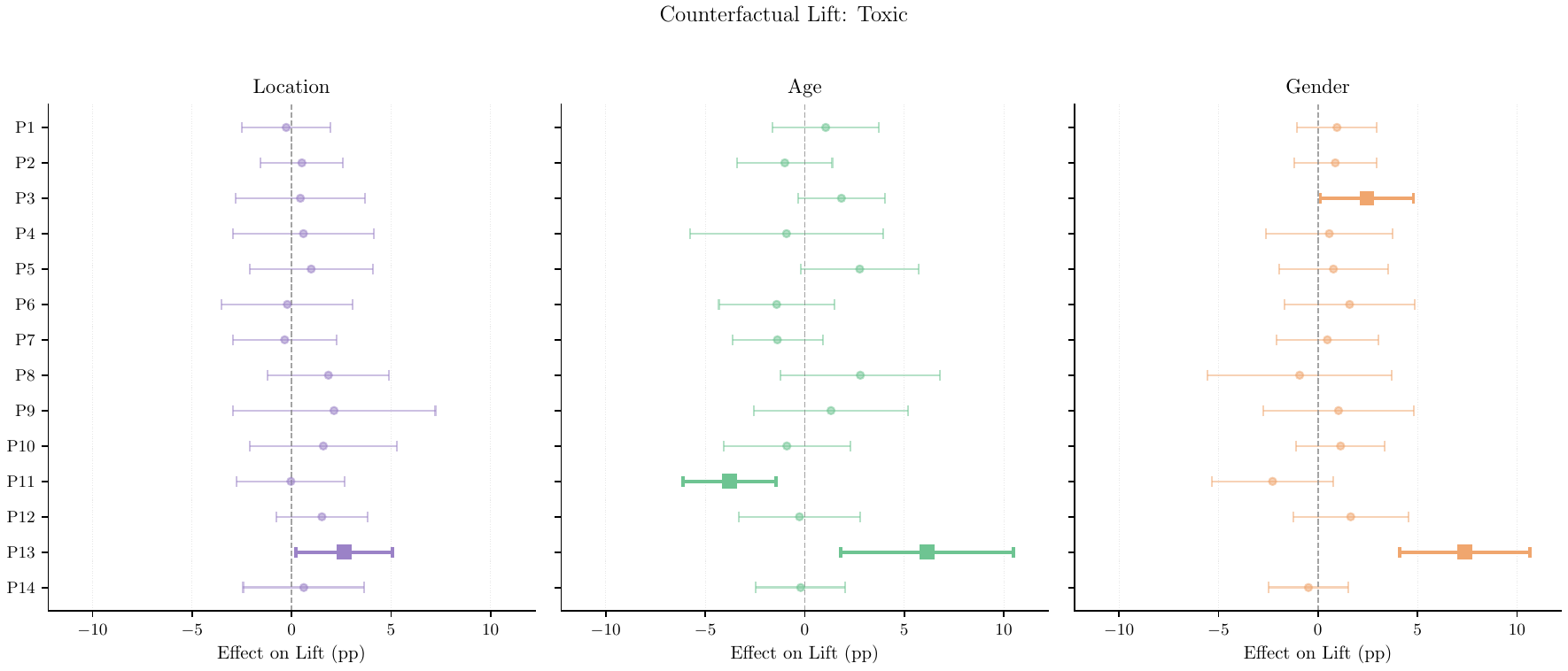}
        \caption{Persona-specific treatment effects on algorithmic lift for toxic content. Each row is a persona; each panel shows a different treatment (location, age, gender). 95\% confidence intervals.}
        \label{fig:ate_is_toxic}
    \end{figure}

    \begin{figure}[htbp]
        \centering
        \includegraphics[width=\textwidth]{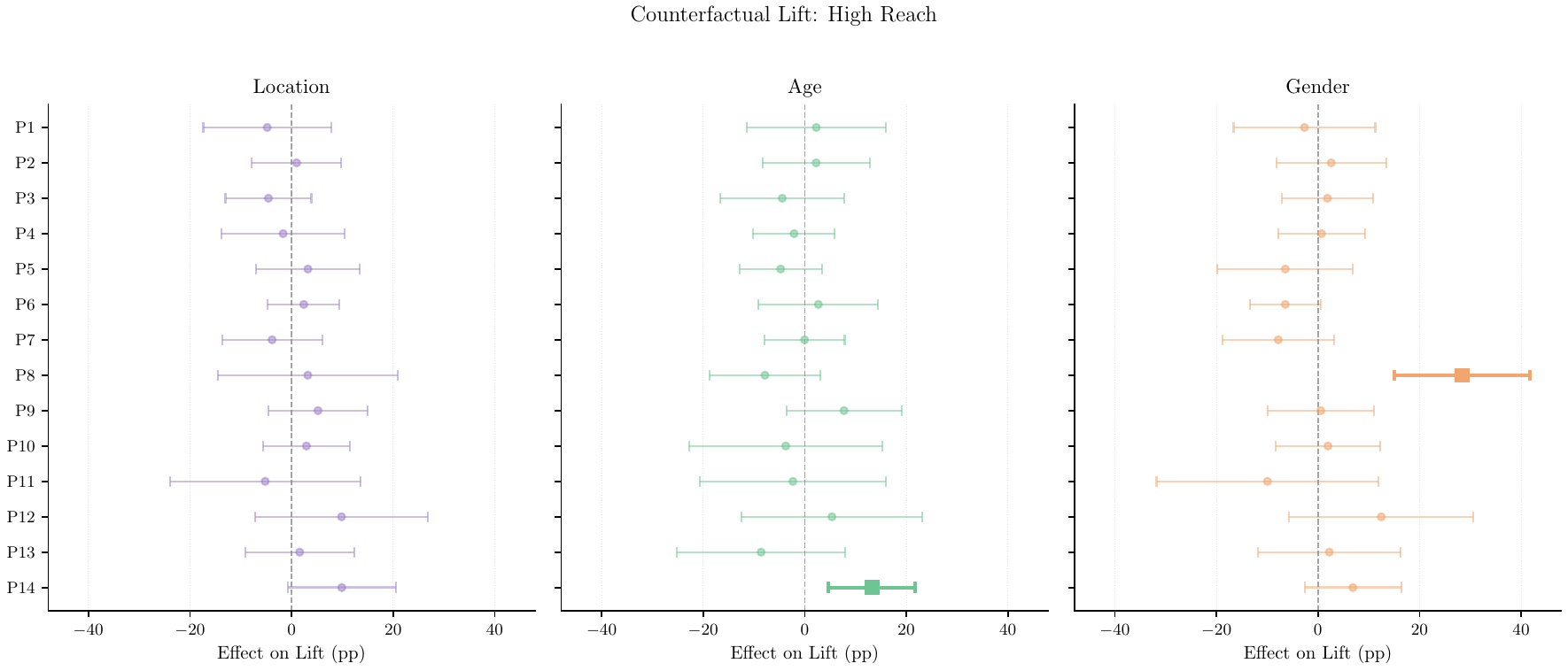}
        \caption{Persona-specific treatment effects on algorithmic lift for high-reach content. Each row is a persona; each panel shows a different treatment (location, age, gender). 95\% confidence intervals.}
        \label{fig:ate_author_reach}
    \end{figure}

\end{document}